\documentclass[10pt,twocolumn,letterpaper]{article}

\usepackage{cvpr}              % To produce the CAMERA-READY version
\usepackage[accsupp]{axessibility}
\usepackage[dvipsnames]{xcolor}

\usepackage{graphicx}
\definecolor{cvprblue}{rgb}{0.21,0.49,0.74}
\usepackage[pagebackref,breaklinks,colorlinks,citecolor=cvprblue]{hyperref}

\usepackage[utf8]{inputenc}
\usepackage{pgfplots}
\pgfplotsset{compat=1.14}
\usepackage{multirow}
\usepackage{float}
\usepackage{subcaption}
\newsavebox{\bigimage}
\usepackage[export]{adjustbox}
\usepackage{tikz}
\usepackage{indentfirst}
\usepackage[ruled,lined,linesnumbered]{algorithm2e}
\DontPrintSemicolon
\usepackage{appendix}
\usepackage{stfloats}
\def\paperID{13968} % *** Enter the Paper ID here
\def\confName{CVPR}
\def\confYear{2024}

\def\modelname{CSTA}
\def\datasetfirst{SumMe}
\def\datasetsecond{TVSum}

\title{CSTA: CNN-based Spatiotemporal Attention for Video Summarization}

\author{Jaewon Son, Jaehun Park, Kwangsu Kim\thanks{Corresponding author}\\
Sungkyunkwan University\\
{\tt\small \{31z522x4,pk9403,kim.kwangsu\}@skku.edu}
}

\begin{document}
\maketitle
% \nolinenumbers
\begin{abstract}
Video summarization aims to generate a concise representation of a video, capturing its essential content and key moments while reducing its overall length. Although several methods employ attention mechanisms to handle long-term dependencies, they often fail to capture the visual significance inherent in frames. To address this limitation, we propose a \textbf{C}NN-based \textbf{S}patio\textbf{T}emporal \textbf{A}ttention (\modelname) method that stacks each feature of frames from a single video to form image-like frame representations and applies 2D CNN to these frame features. Our methodology relies on CNN to comprehend the inter and intra-frame relations and to find crucial attributes in videos by exploiting its ability to learn absolute positions within images. In contrast to previous work compromising efficiency by designing additional modules to focus on spatial importance, \modelname\, requires minimal computational overhead as it uses CNN as a sliding window. Extensive experiments on two benchmark datasets (\datasetfirst\, and \datasetsecond) demonstrate that our proposed approach achieves state-of-the-art performance with fewer MACs compared to previous methods. Codes are available at \url{https://github.com/thswodnjs3/CSTA}.
\end{abstract}
\section{Introduction}
\vspace{-\parskip}
\label{sec:intro}
The rise of social media platforms has resulted in a tremendous surge in daily video data production. Due to the high volume, diversity, or redundancy, it is time-consuming and equally difficult to retrieve the desired content or edit multiple videos. Video summarization is a powerful time-saving technique to condense long videos by retaining the most relevant information, making it easier for users to quickly grasp the main points of the video without having to watch the entire footage.

% The rise of social media platforms has resulted in a tremendous surge in daily video data production. Due to the high volume, diversity, or redundancy, it is time-consuming and equally difficult to retrieve the desired content or edit multiple videos. Video summarization is a powerful time-saving technique to condense long videos by retaining the most relevant information, making it easier for users to quickly grasp the main points of the video without having.

\begin{figure}% >>>
  \centering
  \begin{subfigure}{0.29\columnwidth}
    \centering
    \captionsetup{justification=centering}
    \includegraphics[width=\textwidth]{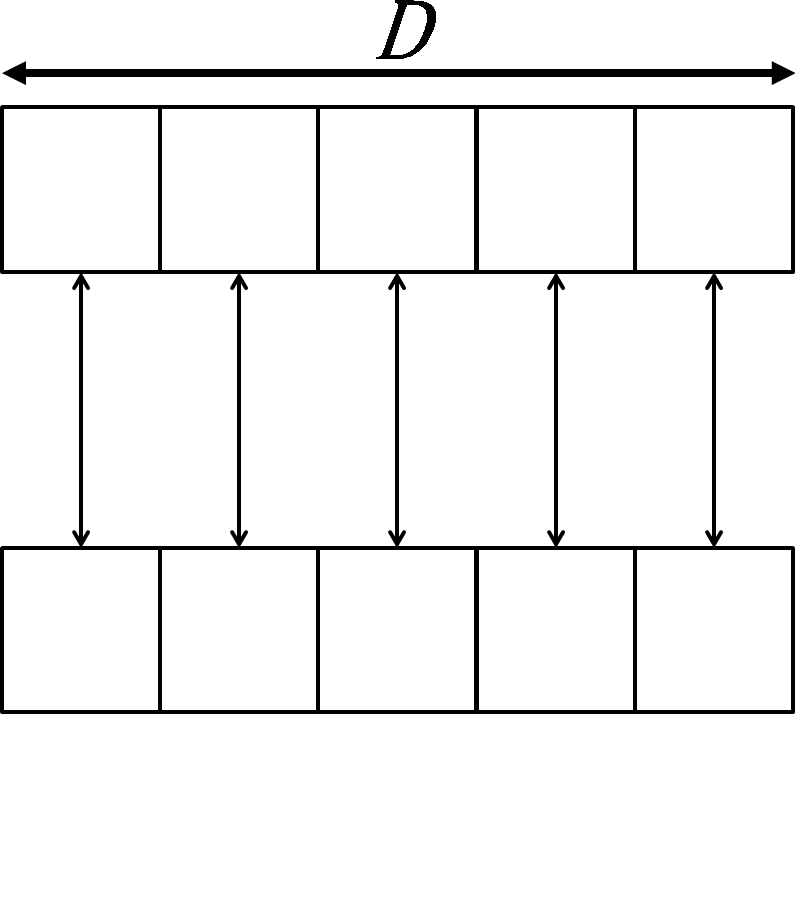}
    \caption{Temporal\\attention}
    \label{fig:figure1-a}
  \end{subfigure}
  \hspace{0.5em}
  \begin{subfigure}{0.29\columnwidth}
    \centering
    \captionsetup{justification=centering}
    \includegraphics[width=\textwidth]{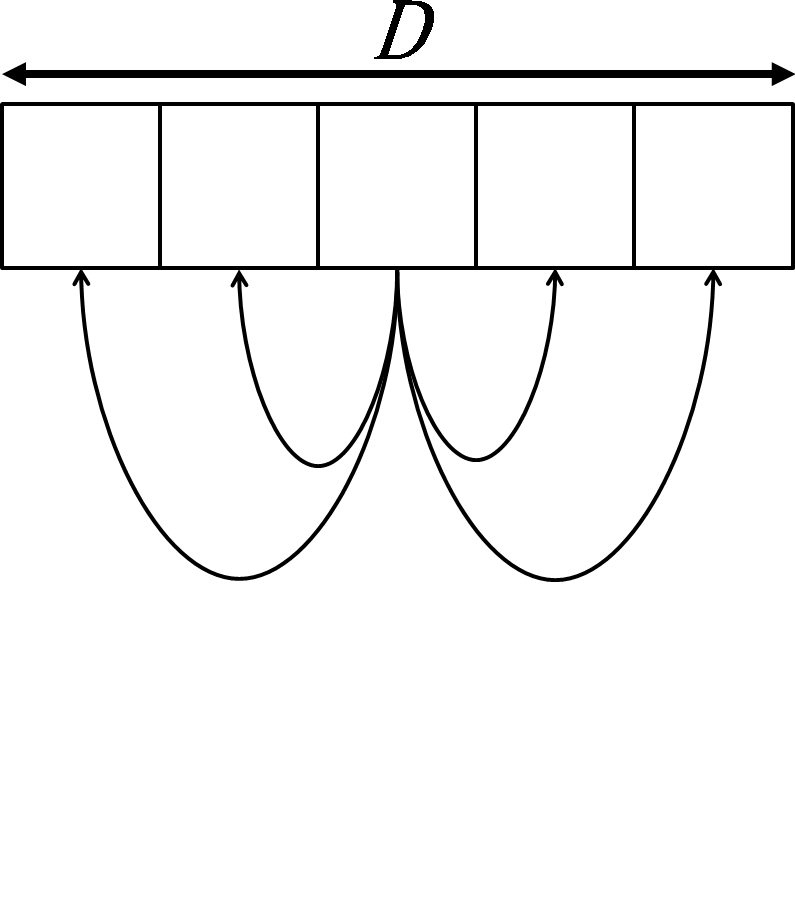}
    \caption{Spatial\\attention}
    \label{fig:figure1-b}
  \end{subfigure}
  \hspace{0.5em}
  \begin{subfigure}{0.331209\columnwidth}
    \centering
    \captionsetup{justification=centering}
    \includegraphics[width=\textwidth]{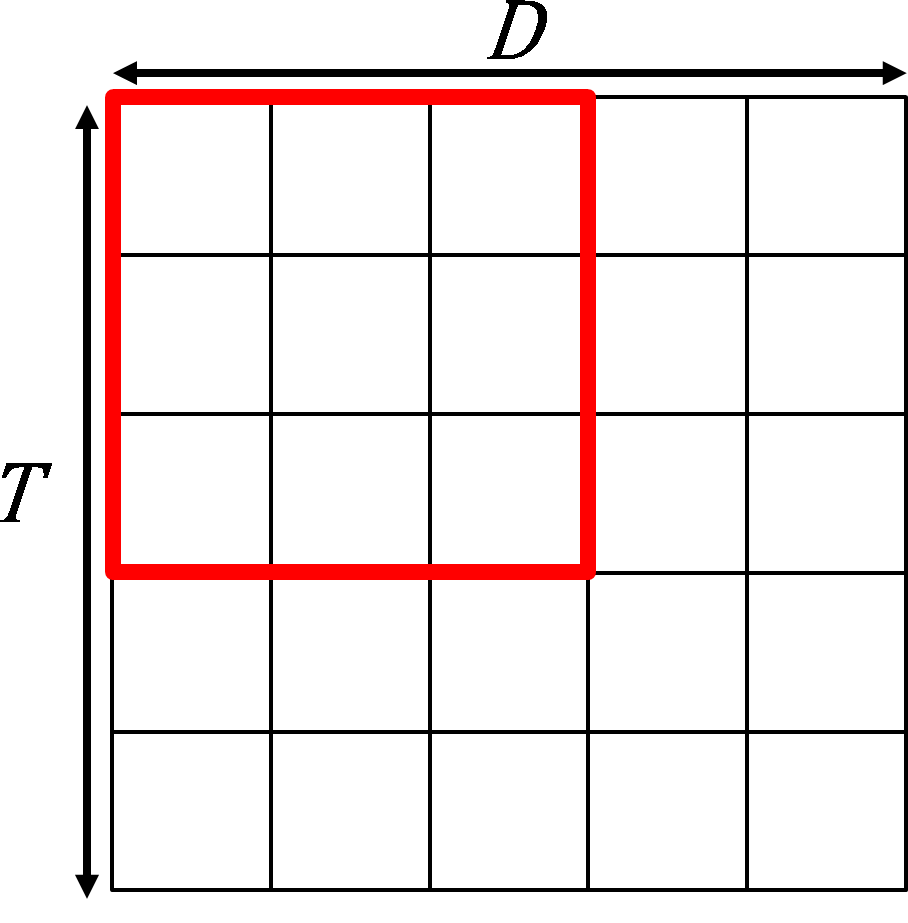}
    \caption{CNN\\attention}
    \label{fig:figure1-c}
  \end{subfigure}
  \caption
    {%
      Approaches for calculating attention. Each row is the feature vector of a frame. \textit{T} is the number of frames, and \textit{D} is the dimension of the feature.%
    }%
  \label{fig:figure1}
\end{figure}%

One of the challenges that occur during video summarization is the long-term dependency problem, where the initial information is often lost due to large data intervals \cite{H_RNN, ADSum, SUM_GDA, VSS_Net}. The decay of initial data prevents deep learning models from capturing the relation between frames essential for determining key moments in videos. Attention \cite{attention}, in which entire frames are reflected through pairwise operations, has gained popularity as a widely adopted technique for solving this problem \cite{M_AVS, VASNet, HMT, STVT, pglsum}. Attention-based models distinguish important parts from unimportant ones by determining the mutual reliance between frames. However, attention cannot consider spatial contexts within images \cite{DMASum, VSS_Net, DAN, STVT, RR_STG}. For instance, current attention calculates temporal attention based on correlations of visual attributes from other frames (See \Cref{fig:figure1-a}), but the importance of visual elements within the frame remains unequal to the temporal significance. Including spatial dependency leads to different weighted values of features, causing changes in temporal importance. Therefore, attention can be calculated more precisely by including visual associations, as shown in \Cref{fig:figure1-b}.

% Prior studies have incorporated spatial importance and demonstrated better performance than solely relying on sequential connections \cite{DMASum, VSS_Net, DAN, STVT, RR_STG}. Nevertheless, acquiring spatial and temporal importance requires the design of additional modules and, thus, incurs excessive costs. Some studies used additional structures to embrace visual relativities in individual frames, such as self-attention \cite{DMASum, STVT}, multi-head attention \cite{VSS_Net}, and graph convolutional neural networks \cite{RR_STG}. Processing too many frames of lengthy videos to capture the temporal and visual importance can be expensive. Thus, obtaining both inter and intra-frame relationships with few computation resources becomes a non-trivial problem.

Prior studies mixed spatial importance and performed better than solely relying on sequential connections \cite{DMASum, VSS_Net, DAN, STVT, RR_STG}. Nevertheless, acquiring spatial and temporal importance requires the design of additional modules and, thus, incurs excessive costs. Some studies used additional structures to embrace visual relativities in individual frames, such as self-attention \cite{DMASum, STVT}, multi-head attention \cite{VSS_Net}, and graph convolutional neural networks \cite{RR_STG}. Processing too many frames of lengthy videos to capture the temporal and visual importance can be expensive. Thus, obtaining both inter and intra-frame relationships with few computation resources becomes a non-trivial problem.

\begin{figure*}
  \centering
  \includegraphics[width=\textwidth]{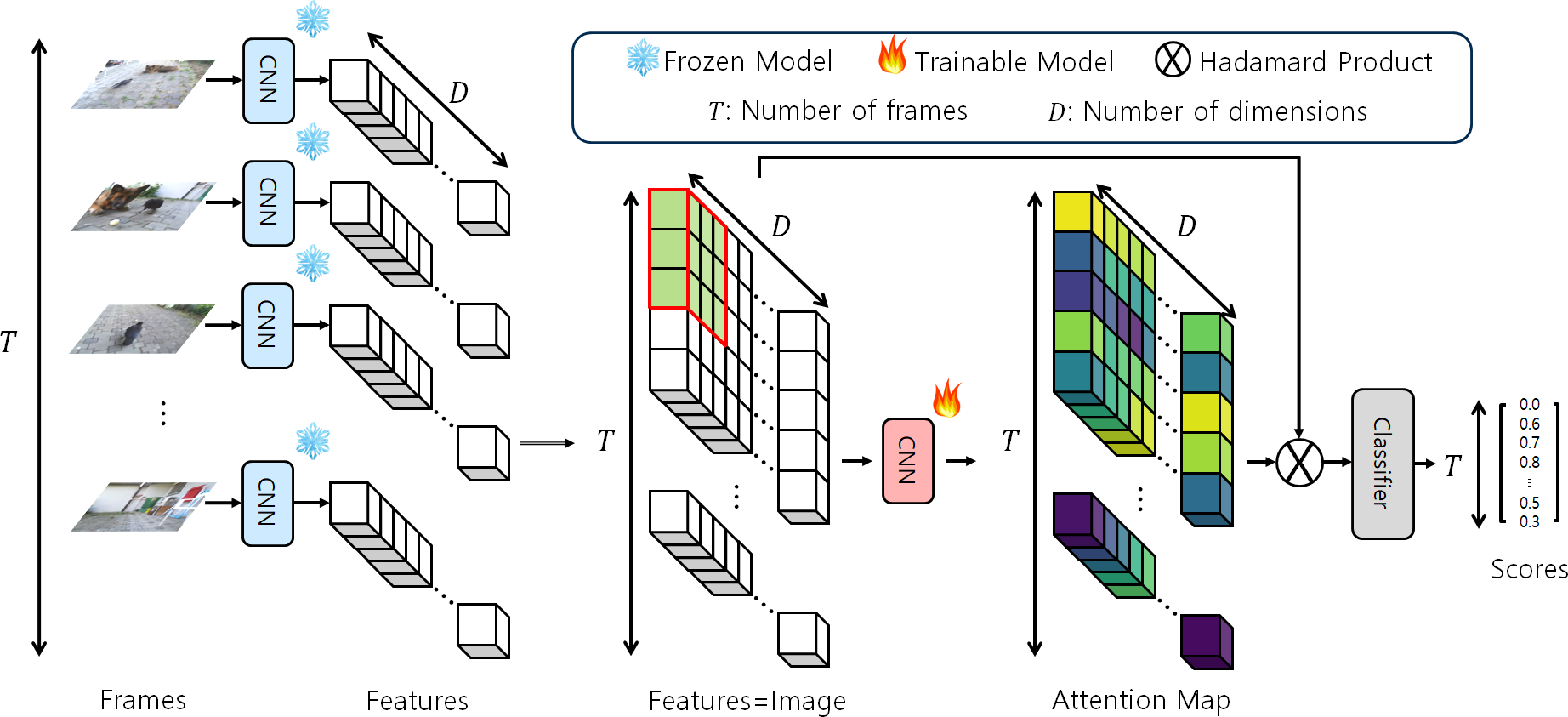}
  \caption{Workflow of \modelname}
  \label{fig:figure2}
\end{figure*}

This paper introduces \textbf{C}NN-based \textbf{S}patio\textbf{T}emporal \textbf{A}ttention (\modelname) to simultaneously capture the visual and ordering reliance in video frames, as shown in \Cref{fig:figure2}. \modelname\, works as follows: Firstly, it extracts features of frames from a video and then concatenates them. Secondly, it treats the assembled frame representations as an image and applies a 2D convolutional neural network (CNN) model to them for producing attention maps. Finally, it combines the attention maps with frame features to predict the importance scores of frames. \modelname\, derives spatial and temporal relationships in the same manner as CNN derives patterns from images, as shown in \Cref{fig:figure1-c}. Further, it searches for vital components in frame representations with the capacity of a CNN to infer absolute positions from images \cite{PosENet, CNN_abs_boundaryeffect}. Unlike previous methods, \modelname\, is efficient as a one-way spatiotemporal processing algorithm because it uses a CNN as a sliding window.

We test the efficacy of \modelname\, on two benchmark datasets - \datasetfirst\,\cite{SumMe} and \datasetsecond\,\cite{TVSum}. Our experiment validates that a CNN produces attention maps from frame features. Further, \modelname\, needs fewer multiply-accumulate operations (MACs) than previous methods for considering the visual and sequential dependency. Our contributions are summarized below:

\begin{itemize}
\item To the best of our knowledge, the proposed model appears to be the first to apply 2D CNN to frame representations in video summarization.
\item The \modelname\, design reflects spatial and temporal associations in videos without requiring considerable computational resources.
\item \modelname\, demonstrates state-of-the-art based on the overall results of two benchmark datasets, \datasetfirst\, and \datasetsecond.
\end{itemize}
\mbox{}
\section{Related Work}
\raggedbottom
\label{sec:related_work}
\subsection{Attention-based Video Summarization}

Many video summarization models use attention to deduce the correct relations between frames and find crucial frames in videos. A-AVS and M-AVS \cite{M_AVS} are encoder-decoder structures in which attention is used to find essential frames. VASNet \cite{VASNet} is based on plain self-attention for better efficiency than encoder-decoder-based ones. SUM-GDA \cite{SUM_GDA} also employs attention for efficiency and supplements diversity into the attention mechanism for generated summaries. CA-SUM \cite{CA_SUM} further enhances SUM-GDA by introducing uniqueness into the attention algorithm in unsupervised ways. Attention in DSNet \cite{DSNet} helps predict scores and precise localization of shots in videos. PGL-SUM \cite{pglsum} has a mechanism to alleviate long-term dependency problems by discovering local and global relationships by applying multi-head attention to segments and the entire video. GL-RPE \cite{GL_RPE} approaches similarly in unsupervised ways by local and global sampling in addition to relative position and attention. VJMHT \cite{VJMHT} uses transformers and improves summarization by learning similarities between analogous videos. CLIP-It \cite{CLIPIt} also relies on the transformers to predict scores by cross-attention between frames and captions of the video. Attention helps models recognize the relations between frames, however, it does not focus on visual relations.

% Many video summarization models use attention to deduce the correct relations between frames and find vital frames in videos. A-AVS and M-AVS \cite{M_AVS} are encoder-decoder structures in which attention is used to find key frames. VASNet \cite{VASNet} is based on plain self-attention for better efficiency than encoder-decoder-based ones. SUM-GDA \cite{SUM_GDA} also uses attention for efficiency and supplements diversity into the attention mechanism for generated summaries. CA-SUM \cite{CA_SUM} enhances SUM-GDA by introducing uniqueness into the attention algorithm in unsupervised ways. Attention in DSNet \cite{DSNet} helps infer scores and precise localization of shots in videos. PGL-SUM \cite{pglsum} is proposed to alleviate long-term dependency by finding local and global relationships by applying multi-head attention to segments and the entire video. GL-RPE \cite{GL_RPE} approaches similarly in unsupervised ways by local and global sampling in addition to relative position and attention. VJMHT \cite{VJMHT} uses transformers and improves summarization by learning similarities between analogous videos. CLIP-It \cite{CLIPIt} also relies on the transformers to infer scores by cross-attention between frames and captions of the video. Attention helps models recognize the relations between frames, however, it does not focus on visual relations.

Visual relevance is vital to understanding video content as it influences the expression of temporal dependency. Some studies have proposed additional networks to find frame-wise visual relationships \cite{DMASum, VSS_Net, DAN, STVT}. The models process the temporal dependency and exploit self-attention or multi-head attention for visual relations of every frame. RR-STG \cite{RR_STG} uses graph CNNs to draw spatial associations using graphs. RR-STG creates graphs based on elements from object detection models \cite{faster_rcnn} to capture the spatial relevance. These methods offer increased performance but incur a high computational cost owing to the separate module handling many frames. This paper adopts CNN as a one-way mechanism for more efficient reflecton of the spatiotemporal importance of multiple frames in long videos.

% Visual relevance is vital to understanding video content as it influences the expression of temporal dependency. Some studies have proposed extra networks to find frame-wise visual relationships \cite{DMASum, VSS_Net, DAN, STVT}. The models process the temporal dependency and exploit self-attention or multi-head attention for visual relations of every frame. RR-STG \cite{RR_STG} uses graph CNNs to draw spatial associations using graphs. RR-STG creates graphs based on elements from object detection models \cite{faster_rcnn} to capture the spatial relevance. These methods offer increased performance but incur a high computational cost owing to the separate module handling many frames. This paper adopts CNN as a one-way mechanism for more efficient reflecton of the spatiotemporal importance of multiple frames in long videos.

\subsection{CNN for Efficiency and Absolute Positions}

CNN is usually employed to resolve computation problems in attention. CvT \cite{CvT} uses CNN for token embedding and projection in vision transformers (ViT) \cite{ViT} and requires a few FLOPs. CeiT \cite{CeiT} uses both CNN and transformers and shows better results with fewer parameters and FLOPs. CmT \cite{CmT} applies depth-wise convolutional operations to obtain a better trade-off between accuracy and efficiency for ViT. We exploit CNN to enhance the efficiency of dealing with multiple frames in video summarization.

CNN can be used for attention by learning absolute positions from images. Islam \etal~ \cite{PosENet} proved that features extracted using a CNN contain position signals. They attributed it to padding, and Kayhan and Germert \cite{CNN_abs_boundaryeffect} verified the same under various paddings. CPVT \cite{CPVT} uses this ability to reflect the position information of tokens and to tackle problems in previous positional encodings for ViT. Based on this behavior of CNNs, our proposed method is designed to seek only the necessary elements for video summarization from frame representations by considering frame features as images.
\section{Method}
\raggedbottom
\label{sec:method}
\subsection{Overview}
\label{sec:method_overview}
\begin{figure*}[ht]% >>>
  \centering
  \includegraphics[height=0.57\textheight]{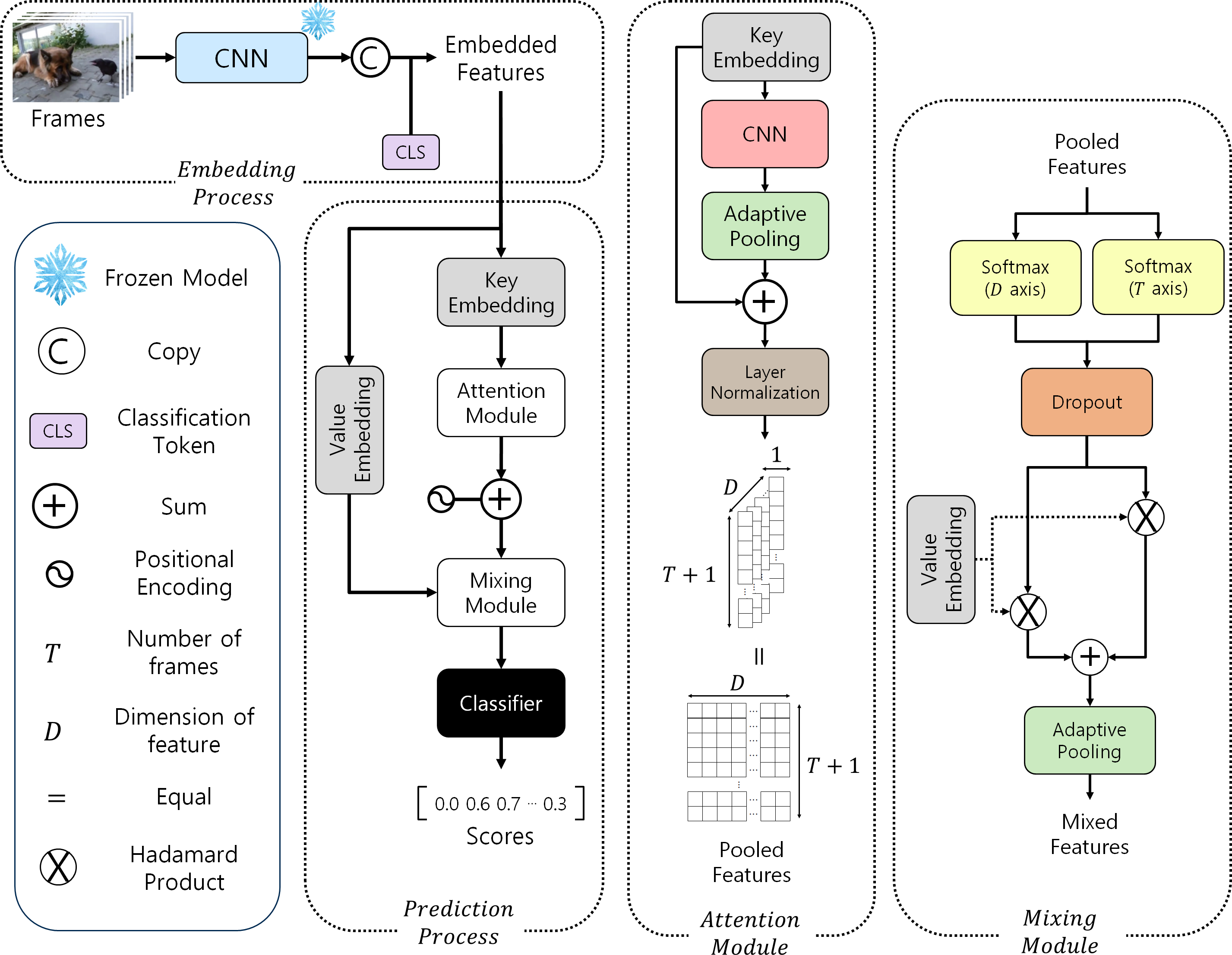}
  \caption{Architecture of \modelname}
  \label{fig:figure3}
\end{figure*}% <<<

This study approaches video summarization as a subset selection problem. We show the proposed \modelname\, framework in \Cref{fig:figure3}. During the \textit{Embedding Process}, the model converts the frames into feature representations. The \textit{Prediction Process} involves using these representations to predict importance scores. In the \textit{Prediction Process}, the \textit{Attention Module} generates attention for videos, and the \textit{Mixing Module} fuses this attention with input frame features. Finally, the \modelname\,  predicts every frame’s importance score, representing the probability of whether the frame should be included in the summary videos. The model is trained by comparing estimated scores and human-annotated scores. During inference, it selects frames based on the knapsack algorithm and creates summary videos using them.

\subsection{Embedding Process}
\label{sec:method_embedding}
\modelname\, converts frames to features for input into the model, as depicted in the \textit{Embedding Process} (\Cref{fig:figure3}). Let the frames be $\mathit{X=\{x_i\}_{t=1}^T}$ when there are $\mathit{T}$ frames in a video, with $\mathit{H}$ as the height and $\mathit{W}$ as the width. Following \cite{vsLSTM,VASNet,DSNet,DMASum,MSVA,SSPVS} for a fair comparison, the frozen pre-trained CNN model (GoogleNet \cite{googlenet}) modifies $\mathit{X \in \mathbb{R}^{T \times 3 \times H \times W}}$ into $\mathit{X' \in \mathbb{R}^{T \times D}}$ where $\mathit{D}$ is the dimension of frame features.\par

To fully utilize the CNN, we replicate the frame representations to match the number of channels (\ie, three). A CNN is usually trained using RGB images \cite{mobilenet, efficientnet, googlenet, resnet}; therefore, pre-trained models are well-optimized on images with three channels. Additionally, we concatenate the classification token (CLS token) \cite{ViT, STVT} into frame features:\par

\begin{equation}
  \mathit{X'' = Concat_{axis=0}(X',X',X')}
  \label{eq:equation1}
\end{equation}

\begin{equation}
  \mathit{E = Concat_{axis=1}(X_{CLS},X'')}
  \label{eq:equation2}
\end{equation}

where $\mathit{X'' \in \mathbb{R}^{3 \times T \times D}}$ and $\mathit{X_{CLS} \in \mathbb{R}^{3 \times 1 \times D}}$ are the appended feature and the CLS token, respectively. $\mathit{E \in \mathbb{R}^{3 \times (T+1) \times D}}$ is the embedded feature. $\mathit{Concat_{axis=0}}$ and $\mathit{Concat_{axis=1}}$ concatenate features in the channel axis and $\mathit{T}$ axis, respectively. Motivated by STVT \cite{STVT}, we append the CLS token with input frame features. The CLS token is the learnable parameters fed into the models with inputs and trained with models jointly. STVT obtains correlations of frames using the CLS token and aggregates the CLS token with input frames to capture global contexts. We follow the same method in prepending and combining the CLS token with frame features. The fusing process is completed later in the \textit{Mixing Module}.

\subsection{Prediction Process}
\label{sec:method_prediction}
\modelname\, calculates importance scores for $\mathit{T}$ frames, as shown in the \textit{Prediction Process} (\Cref{fig:figure3}). The classifier assigns scores to frames after the \textit{Attention Module} and \textit{Mixing Module}. The \textit{Attention Module} makes attention maps from $\mathit{E}$, and the \textit{Mixing Module} aggregates this attention with $\mathit{E}$. A detailed explanation is given in \Cref{alg:alg1}.

\begin{algorithm}
\caption{\textit{Prediction Process}}\label{alg:alg1}
        \SetKwInOut{Input}{input}
        \SetKwInOut{Output}{output}
        \Input{$E \in \mathbb{R}^{3 \times (T+1) \times D}$}
        \Output{$\mathit{S} \in \mathbb{R}^\mathit{T}$}
    \SetKwBlock{Beginn}{beginn}{ende}
    \Begin{
            $E^K = W^{K}E$\label{alg:alg1_1}\;
            $E^V = W^{V}E[0]$\label{alg:alg1_2}\;
            \;
            $P = \textit{Attention Module}(E^K)$\label{alg:alg1_3}\text{\,\,\,\,\,\,\,\,\,\,\,\,\,\,\,\,\Cref{sec:method_AM}}\;
            % $P = Attention Module(E^K)$\label{alg:alg1_3}\tcp*[l]{\Cref{sec:method_AM}}
            $P_{pos} = P + \textit{Positional Encoding}$\label{alg:alg1_4}\;
            $M = \textit{Mixing Module}(P_{pos}, E^V)$\label{alg:alg1_5}\text{ 
 \,\,\,\,\Cref{sec:method_MM}}\;
            $S = Classifier(M)$\label{alg:alg1_6}\;
        \Return $S$
    }
\end{algorithm}

We generate the key and value from $\mathit{E}$ by using two linear layers based on the original attention \cite{attention}. The metrics $\mathit{W^K}$ and $\mathit{W^V \in \mathbb{R}^{D \times D}}$ are weights of linear layers projecting $\mathit{E}$ into the key and value (\cref{alg:alg1_1}-\cref{alg:alg1_2}). Unlike $\mathit{E^K}$, \modelname\, uses a single channel of frame features in $\mathit{E}$ to produce features by value embedding (\cref{alg:alg1_2}) because we only need one $\mathit{X'}$ except for duplicated ones, which are simply used for reproducing image-like features. We select the first index as a representative, which is $\mathit{E[0]} \in \mathbb{R^{\mathit{(T+1) \times D}}}$.\par

The \textit{Attention Module} processes spatiotemporal characteristics and focuses on critical attributes in $\mathit{E^K}$ (\cref{alg:alg1_3}). We add positional encodings to $\mathit{P}$ to strengthen the absolute position awareness further (\cref{alg:alg1_4}). Unlike the prevalent way of adding positional encoding into inputs \cite{attention, ViT}, this study adds positional encoding into the attention maps based on \cite{pglsum}. This is because adding positional encodings into input features distorts images so that models can recognize this distortion as different images. Moreover, models cannot fully recognize these absolute position encodings in images during training owing to a lack of data. Therefore, \modelname\, makes $\mathit{P_{pos}}$ by attaching positional encodings to attentive features $\mathit{P}$.\par

The \textit{Mixing Module} inputs $\mathit{P_{pos}}$ and $\mathit{E^V}$ and produces mixed features $\mathit{M} \in \mathbb{R}^{(T+1) \times D}$ (\Cref{alg:alg1_5}). The classifier predicts importance scores vectors $\mathit{S} \in \mathbb{R}^\mathit{T}$ from $\mathit{M}$ (\Cref{alg:alg1_6}).

\subsection{Attention Module}
\label{sec:method_AM}
The \textit{Attention Module} (\Cref{fig:figure3}) produces attention maps by utilizing a trainable CNN (GoogleNet \cite{googlenet}) handling frame features $\mathit{E^K}$. The CNN captures the spatiotemporal dependency using kernels, similar to how a CNN learns from images, as shown in \Cref{fig:figure1-c}. The CNN also searches for essential elements from $\mathit{E^K}$ for summarization, with the ability to learn absolute positions. Based on \cite{PosENet,CNN_abs_boundaryeffect, CPVT}, CNN imbues representations with positional information so that \modelname\, can encode the locations of significant attributes from frame features for summarization.\par

We make the shape of attention maps the same as that of input features to aggregate attention maps with input features. This study leverages two strategies for equal scale: deploying the adaptive pooling operation and using the same CNN model (GoogleNet \cite{googlenet}) in the \textit{Embedding Process} and \textit{Attention Module}. Pooling layers reduce the scale of features in the CNN; therefore, the size of outputs from the CNN is changed from $\mathit{E^K} \in \mathbb{R}^{3 \times (T+1) \times D}$ to $\mathit{E^K_{CNN}} \in \mathbb{R}^{D \times \frac{T+1}{r} \times \frac{D}{r}}$, where $\mathit{r}$ is the reduction ratio. To expand diverse lengths of frame representations, we exploit adaptive pooling layers to adjust the shape of features by bilinear interpolation. Furthermore, the number of output channels from the learnable CNN equals the dimension of frame features from the fixed CNN because of the same CNN models. The output from adaptive pooling is $\mathit{E^K_{pool}} \in \mathbb{R}^{D \times (T+1) \times 1}$.\par

As suggested in \cite{resnet}, this study uses a skip connection:

\begin{equation}
  \mathit{P = LayerNorm(E^K_{pool} + E^K[0])}
  \label{eq:equation3}
\end{equation}

where the output is $\mathit{P} \in \mathbb{R}^{D \times (T+1)}$, followed by layer normalization \cite{layernorm}. A skip connection supports more precise attention and stable training in \modelname. As same with $\mathit{E^V}$, explained in \Cref{alg:alg1} (\Cref{alg:alg1_2}), we only use the single frame feature of $\mathit{E^K}$ and ignore replications of frame features.\par

The size of $\mathit{P}$ is equal to the size of frame features with $\mathit{(T+1) \times D}$; therefore, each value of $\mathit{P}$ has the spatiotemporal importance of frame features. By combining $\mathit{P}$ with frame features, the \modelname\, reflects the sequential and visual significance of frames. After supplementing the positional encodings, $\mathit{P_{pos}}$ will be used as inputs for the \textit{Mixing Module}.

\subsection{Mixing Module}
\label{sec:method_MM}
In the \textit{Mixing Module} (\Cref{fig:figure3}), we employ softmax along the time and dimension axes to compute the temporal and visual weighted values of $\mathit{P_{pos}}$:
\begin{equation}
  \mathit{Att_T: \sigma(d_{i}) = \left(\frac{e^{d_{i}}}{ \sum\limits_{j} e^{d_{j}}}\right)j = 1,...,T+1}
  \label{eq:equation4}
\end{equation}

\begin{equation}
  \mathit{Att_D: \sigma(d_{i}) = \left(\frac{e^{d_{i}}}{ \sum\limits_{k} e^{d_{k}}}\right)k = 1,...,D}
  \label{eq:equation5}
\end{equation}

where $\mathit{Att_T}$ is the temporal importance, and $\mathit{Att_D}$ is the visual importance. \Cref{eq:equation4} calculates the weighted values between $\mathit{T+1}$ frames, including the CLS token, in the same dimension. \Cref{eq:equation5} computes the weighted values between different dimensions in the same frame. $\mathit{Att_D}$ represents the spatial importance because each value of the dimension from features includes visual characteristics by CNN, processing image patterns, and producing informative vectors.

After acquiring weighted values, a dropout is employed for these values before integrating them with $\mathit{E^V}$. The dropout erases parts of features by setting 0 values for better generalization; it also works for attention, as shown in \cite{attention,pglsum}. If a dropout is applied to inputs as in the original attention \cite{attention}, the CNN cannot learn contexts from 0 values, unlike self-attention, because the dropout spoils the local contexts of deleted parts. Therefore, we follow \cite{pglsum} by applying the dropout to the output of the softmax operations for generalization.\par

After dropout, the \modelname\, combines the spatial and temporal importance with the frame features:

\begin{equation}
  \mathit{M = Att_T \odot E^V + Att_D \odot E^V}
  \label{eq:equation6}
\end{equation}

where $\odot$ is the element-wise multiplication, and $\mathit{M \in \mathbb{R^{\mathit{(T+1) \times D}}}}$ is the mixed representations. \modelname\, reflects weighted values into frame features by blending $\mathit{Att_T}$ and $\mathit{Att_D}$ with $\mathit{E^V}$ by element-wise multiplication. Incorporating visual and sequential attention values by addition encompasses spatiotemporal importance at the same time.\par

Subsequently, to integrate the CLS token with frame features, adaptive pooling transforms $\mathit{M \in \mathbb{R}^{(T+1) \times D}}$ into $\mathit{M' \in \mathbb{R}^{T \times D}}$ by average. Unlike STVT \cite{STVT}, in which linear layers are used to merge the CLS token with constant numbers of frames, \modelname\, uses adaptive pooling to cope with various lengths of videos. Adaptive pooling fuses the CLS token with a few frames; however, it intensifies our model owing to the generalization of the classifier, which consists of fully connected layers. $\mathit{M'}$ from adaptive pooling enters into the classifier computing importance scores of frames.

% Subsequently, adaptive pooling reshapes $\mathit{M \in \mathbb{R}^{(T+1) \times D}}$ into $\mathit{M' \in \mathbb{R}^{T \times D}}$ to integrate the CLS token with frame features by average. Unlike STVT \cite{STVT}, in which linear layers are used to merge the CLS token with constant numbers of frames, \modelname\, uses adaptive pooling to cope with various lengths of videos. Adaptive pooling fuses the CLS token with a few frames; however, it intensifies our model owing to the generalization of the classifier, which consists of fully connected layers. $\mathit{M'}$ from adaptive pooling enters into the classifier computing importance scores of frames.

\subsection{Classifier}
\label{sec:method_inference}
Based on the output of the adaptive pooling, the classifier exports the importance scores. We follow \cite{VASNet, pglsum, A2Summ} to construct the structure of the classifier as follows:
\begin{equation}
  \mathit{R = LayerNorm(Dropout(ReLU(FC(M'))))}
  \label{eq:equation7}
\end{equation}

\begin{equation}
  \mathit{S = Sigmoid(FC(R))}
  \label{eq:equation8}
\end{equation}

where $\mathit{R \in \mathbb{R}^{T \times D}}$ is derived after $\mathit{M'}$ passes through a fully connected layer, relu, dropout, and layer normalization. Another fully connected layer maps the representation of each frame into single values, and the sigmoid computes scores $\mathit{S \in \mathbb{R}^{T}}$.\par

We train \modelname\, by comparing predicted and ground truth scores. For the loss function, we use the mean squared loss as follows:

\begin{equation}
  \mathit{Loss = \frac{1}{T}\sum (S_p - S_g)^2}
  \label{eq:equation9}
\end{equation}

where $\mathit{S_p}$ is the predicted score, and $\mathit{S_g}$ is the ground truth score.\par

The \modelname\, creates summary videos based on shots that KTS \cite{KTS} derives. It computes the average importance scores of shots into which KTS splits videos \cite{vsLSTM}. The summary videos consist of shots with two constraints:\par

\begin{equation}
  \mathit{max \sum S_i}
  \label{eq:equation10}
\end{equation}

\begin{equation}
  \mathit{\sum Length_i \leq 15\%}
  \label{eq:equation11}
\end{equation}

where $\mathit{i}$ is the index of selected shots. $\mathit{S_i \in [0, 1]}$ is the importance score of the $\mathit{i}$th shot between 0 and 1, and $\mathit{Length_i}$ is the percentage of the length of the $\mathit{i}$th shot in the original videos. Our model picks shots with high scores by exploiting the 0/1 knapsack algorithm as in \cite{TVSum}. Following \cite{SumMe}, summary videos have a length limit of 15\% of the original videos.
\section{Experiments}
\raggedbottom
\label{sec:experiment}
\subsection{Settings}
\label{sec:exp_setting}
\paragraph{Evaluation Methods.} 
We evaluate \modelname\, using Kendall’s ($\mathit{\tau}$) \cite{kendall} and Spearman’s ($\mathit{\rho}$) \cite{spearman} coefficients. Both metrics are rank-based correlation coefficients that are used to measure the similarities between model-estimated and ground truth scores. The F1 score is the most commonly used metric in video summarization; however, it has a significant drawback when used to evaluate summary videos. Based on \cite{problem_fscore, MAAM}, due to the limitation of the summary length, the F1 score is evaluated to be higher if models choose as many short shots as possible and ignore long key shots. This fact implies that the F1 score might not represent the correct performance in video summarization. A detailed explanation of how to measure correlations is provided in \Cref{sec:sup_measure_correlation}.

\paragraph{Datasets.}
This study utilizes two standard video summarization datasets - \datasetfirst\,\cite{SumMe} and \datasetsecond\,\cite{TVSum}. \datasetfirst\, consists of videos with different contents (\eg, holidays, events, sports) and various types of camera angles (\eg, static, egocentric, or moving cameras). The videos are raw or edited public ones with lengths of 1-6 minutes. At least 15 people create ground truth summary videos for all data, and the models predict the average number of selections by people for every frame.  \datasetsecond\, comprises 50 videos from 10 genres (\eg, documentaries, news, vlogs). The videos are 2-10 minutes long, and 20 people annotated the ground truth for each video. The ground truth is a shot-level importance score ranging from 1 to 5, and models try to estimate the average shot-level scores.
\\
\\
\textbf{Implementation details} are explained in \Cref{sec:sup_implementation_details}.

\subsection{Verification of Attention Maps being Created using  CNN}
\label{exp_CNN}

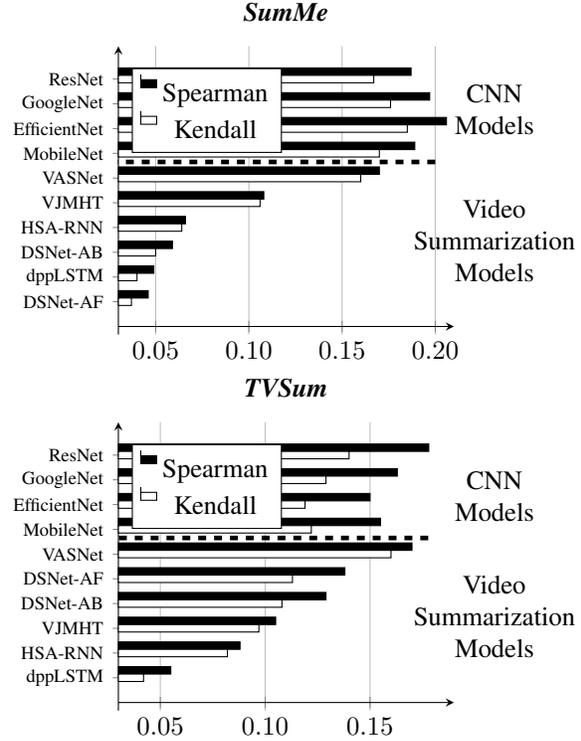
\begin{figure}
  \raggedleft
  \begin{tikzpicture}
  \node[align=center] at (5,3) {CNN\\Models};
  \node[align=center] at (5,1.25) {Video\\Summarization\\Models};
  \begin{axis}[
    width=0.725\columnwidth,height=0.65\columnwidth,
    axis lines=middle,title={\textit{\textbf{\datasetfirst}}},
    enlarge y limits={abs=0.025\textwidth},
    yticklabel style={anchor=east,font=\scriptsize},
    yticklabels from table={comp_summe.dat}{Model},ytick=data,
    xticklabel style={/pgf/number format/fixed, /pgf/number format/fixed zerofill, /pgf/number format/precision=2},
    xmin=0.03,xmax=0.21,xtick={0.05,0.10,0.15,0.20},xmajorgrids,
    ybar legend,
    legend image post style={scale=1},
    legend style={nodes={scale=1},at={(0.04,0.63)},anchor=south west}
    ]
  \draw [dashed,line width=1.75pt] (0,6.65) -- (0.2,6.65);
  \addplot[
    xbar,
    bar width=.3,
    bar shift=+0.3,
    fill=black,
  ] table [x=Spearman,y=X]{comp_summe.dat};
  \addplot[
    xbar,
    bar width=.3,
    fill=white,
  ] table [x=Kendall,y=X]{comp_summe.dat};
  \legend{Spearman,Kendall}
  \end{axis}
  \end{tikzpicture}
  \raggedright
  \begin{tikzpicture}
  \node[align=center] at (5,2.85) {CNN\\Models};
  \node[align=center] at (5,1.25) {Video\\Summarization\\Models};
  \begin{axis}[
    width=0.725\columnwidth,height=0.65\columnwidth,
    axis lines=middle,title={\textit{\textbf{\datasetsecond}}},
    enlarge y limits={abs=0.025\textwidth},
    yticklabel style={anchor=east,font=\scriptsize},
    yticklabels from table={comp_tvsum.dat}{Model},ytick=data,
    xticklabel style={/pgf/number format/fixed, /pgf/number format/fixed zerofill, /pgf/number format/precision=2},
    xmin=0.03,xmax=0.19,xtick={0.05,0.10,0.15},xmajorgrids,
    ybar legend,
    legend image post style={scale=1},
    legend style={nodes={scale=1},at={(0.04,0.63)},anchor=south west}
    ]
  \draw [dashed,line width=1.75pt] (0,6.65) -- (0.18,6.65);
  \addplot[
    xbar,
    bar width=.3,
    bar shift=+0.3,
    fill=black,
  ] table [x=Spearman,y=X]{comp_tvsum.dat};
  \addplot[
    xbar,
    bar width=.3,
    fill=white,
  ] table [x=Kendall,y=X]{comp_tvsum.dat};
  \legend{Spearman,Kendall}
  \end{axis}
  \end{tikzpicture}
  \caption{Comparison of summarizing performance between CNN and video summarization models. The x-axis shows performance, and the y-axis shows model names. Based on the dashed line, the performance of CNN is displayed above, and the video summarization models are below.}
  \label{fig:figure4}
\end{figure}

Previous studies on video summarization have yet to apply 2D CNN directly to frame features. Therefore, we verify that CNN can create attention maps from frame features. We choose MobileNet-V2 \cite{mobilenet}, EfficientNet-B0 \cite{efficientnet}, GoogleNet \cite{googlenet}, and ResNet-18 \cite{resnet} as CNN models since we focus on limited computation costs. This study applies CNN models to frame features and trains them to compute the frame-level importance scores without the classifier. The CNN directly exports $\mathit{T}$ scores by inputting its output features into the adaptive pooling layer with a target shape $\mathit{T \times 1}$. As the importance score of each frame is between 0 and 1, each score is similar to the weighted value of each frame. Thus, we can test whether CNN generates attention maps based on the video summarization performance. Surprisingly, the CNN models predict the importance scores much better than the previous video summarization models on \datasetfirst\,, as shown in \Cref{fig:figure4}. Even though the CNN models do not perform best on \datasetsecond, they still show promising performance compared to existing video summarization models. The results show that the CNN produces attention maps by capturing the spatiotemporal relations and detecting crucial attributes in frame features based on absolute position encoding ability, unlike conventional methods that solely address the temporal dependency.

\subsection{Performance Comparison}
\label{exp_comparison}
\begin{table}
  \centering
  \resizebox{0.475\textwidth}{!}{
    \begin{tabular}{c|ccc|ccc}
      \noalign{\smallskip}\noalign{\smallskip}
      \hline
      \multirow{2}{*}{Method} & \multicolumn{3}{c|}{\datasetfirst} & \multicolumn{3}{c}{\datasetsecond} \\
      % \cline{1-7}
          & Rank & $\mathit{\tau}$ & $\mathit{\rho}$ & Rank & $\mathit{\tau}$ & $\mathit{\rho}$ \\
      \hline
       Random & - & 0.000 & 0.000 & - & 0.000 & 0.000 \\
       Human & - & 0.205 & 0.213 & - & 0.177 & 0.204 \\
       \midrule
       dppLSTM\cite{vsLSTM} & 15 & 0.040 & 0.049 & 22 & 0.042 & 0.055 \\
       DAC\cite{DAC}$\mathit{^T}$ & 12.5 & 0.063 & 0.059 & 21 & 0.058 & 0.065 \\
       HSA-RNN\cite{HSA_RNN} & 11.5 & 0.064 & 0.066 & 19.5 & 0.082 & 0.088 \\
       DAN\cite{DAN}$\mathit{^{ST}}$ & - & - & - & 19.5 & 0.071 & 0.099 \\
       STVT\cite{STVT}$\mathit{^{ST}}$ & - & - & - & 15.5 & 0.100 & 0.131 \\
       DSNet-AF\cite{DSNet}$\mathit{^T}$ & 16 & 0.037 & 0.046 & 13.5 & 0.113 & 0.138 \\
       DSNet-AB\cite{DSNet}$\mathit{^T}$ & 13.5 & 0.051 & 0.059 & 15 & 0.108 & 0.129 \\
       HMT\cite{HMT}$\mathit{^M}$ & 10.5 & 0.079 & 0.080 & 17.5 & 0.096 & 0.107 \\
       VJMHT\cite{VJMHT}$\mathit{^T}$ & 8.5 & 0.106 & 0.108 & 17.5 & 0.097 & 0.105 \\
       CLIP-It\cite{CLIPIt}$\mathit{^M}$ & - & - & - & 13.5 & 0.108 & 0.147 \\
       iPTNet\cite{iPTNet}$\mathit{^+}$ & 8.5 & 0.101 & 0.119 & 11 & 0.134 & 0.163 \\
       A2Summ\cite{A2Summ}$\mathit{^M}$ & 7 & 0.108 & 0.129 & 10 & 0.137 & 0.165 \\
       VASNet\cite{VASNet}$\mathit{^T}$ & 6 & 0.160 & 0.170 & 9 & 0.160 & 0.170 \\
       AAAM\cite{MAAM}$\mathit{^T}$ & - & - & - & 6.5 & 0.169 & 0.223 \\
       MAAM\cite{MAAM}$\mathit{^T}$ & - & - & - & 5.5 & 0.179 & 0.236 \\
       VSS-Net\cite{VSS_Net}$\mathit{^{ST}}$ & - & - & - & 3 & 0.190 & 0.249 \\
       DMASum\cite{DMASum}$\mathit{^{ST}}$ & 11 & 0.063 & 0.089 & \textbf{1} & \textbf{0.203} & \textbf{0.267} \\
       RR-STG\cite{RR_STG}$\mathit{^{ST}}$ & 2.5 & 0.211* & 0.234 & 7.5 & 0.162 & 0.212 \\
       MSVA\cite{MSVA}$\mathit{^M}$ & 3.5 & 0.200 & 0.230 & 5.5 & 0.190 & 0.210 \\
       SSPVS\cite{SSPVS}$\mathit{^M}$ & 3* & 0.192 & 0.257* & 4.5 & 0.181 & 0.238 \\
       \bottomrule
       GoogleNet\cite{googlenet}$\mathit{^{ST}}$ & 5 & 0.176 & 0.197 & 11.5 & 0.129 & 0.163 \\
       \modelname$\mathit{^{ST}}$ & \textbf{1} & \textbf{0.246} & \textbf{0.274} & 2* & 0.194* & 0.255*
    \end{tabular}
  }
  \caption{Comparison between \modelname\, and state-of-the-art on \datasetfirst\, and \datasetsecond. Rank is the average rank between Kendall's ($\mathit{\tau}$) and Spearman's ($\mathit{\rho}$) coefficients. We categorize different types of video summarization models: temporal ($\mathit{T}$) and spatiotemporal ($\mathit{ST}$) attention-based, multi-modal based ($\mathit{M}$), and external dataset-based ($\mathit{+}$) models. The scores marked in bold and by the asterisk are the best and second-best ones, respectively. GoogleNet is the baseline model. Note that all feature extraction models are CNNs for a fair comparison.}
  \label{tab:tab1}
\end{table}

We compare \modelname\, with existing state-of-the-art methods on \datasetfirst\, and \datasetsecond. The results in \Cref{tab:tab1} show that \modelname\, achieves the best performance on \datasetfirst\, and the second-best score on \datasetsecond\, based on the average rank. DMASum \cite{DMASum} shows the best performance on \datasetsecond\, but does not perform well on \datasetfirst, as indicated in \Cref{tab:tab1}. DMASum has $\mathit{\tau}$ and $\mathit{\rho}$ coefficients of 0.203 and 0.267 on \datasetsecond, respectively, whereas 0.063 and 0.089 on \datasetfirst, respectively. This implies that \modelname\, provides more stable performances than DMASum, although it provides slightly lower performance than DMASum on \datasetsecond. Based on the overall performance of both datasets, our \modelname\, has achieved state-of-the-art results.

Further, \modelname\, excels in video summarization models relying on classical pairwise attention \cite{DAC, DSNet, VJMHT, VASNet, MAAM}, focusing on temporal attention only. This clarifies that considering the visual dependency helps \modelname\, understand crucial moments by capturing meaningful visual contexts. Like \modelname, some approaches, including DMASum, focus on spatial and temporal dependency \cite{DAN, STVT, VSS_Net, RR_STG}, but they perform poorly compared to our proposed methodology. This is because CNN is much more helpful than previous methods by using the ability to learn the absolute position in frame features.

\modelname\, also outperforms methods that require additional datasets from other modalities or tasks \cite{HMT, CLIPIt, iPTNet, A2Summ, MSVA, SSPVS}. Our observations suggest that \modelname\, can find essential moments in videos solely based on images without assistance from extra data. We also show the visualization of generated summary videos from different models in \Cref{sec:sup_visualize_sum}.

\subsection{Ablation Study}
\label{exp_ablation}
\begin{table}
  \raggedleft
  \resizebox{0.475\textwidth}{!}{
    \raggedright
    \begin{tabular}{l|cc|cc}
      \noalign{\smallskip}\noalign{\smallskip}
      \hline
      \multirow{2}{*}{Module} & \multicolumn{2}{c|}{\datasetfirst} & \multicolumn{2}{c}{\datasetsecond} \\
      % \cline{1-5}
         & $\mathit{\tau}$ & $\mathit{\rho}$ & $\mathit{\tau}$ & $\mathit{\rho}$ \\
      \hline
       GoogleNet (Baseline) & 0.176 & 0.197 & 0.129 & 0.163 \\
       $\mathit{(+)}$\textit{Attention Module} & 0.184 & 0.205 & 0.176 & 0.231 \\
       $\mathit{(+)Att_D}$ & 0.189 & 0.211 & 0.182 & 0.240 \\
       $\mathit{(+)}$\textit{Key,Value Embedding} & 0.207 & 0.231 & 0.193 & 0.253 \\
       $\mathit{(+)}$\textit{Positional Encoding} & 0.225 & 0.251 & 0.189 & 0.248 \\
       $\mathit{(+)X_{CLS}}$ & 0.231 & 0.257 & 0.193 & 0.254 \\
       $\mathit{(+)}$\textit{Skip Connection} & 0.246 & 0.274 & 0.194 & 0.255 \\
    \end{tabular}
  }
  \caption{We listed Kendall's ($\mathit{\tau}$) and Spearman's ($\mathit{\rho}$) coefficients for different modules. $\mathit{(+)}$ denotes the stacking of modules on top of the previous ones.}
  \label{tab:tab2}
\end{table}

This study verifies all components step-by-step, as indicated in \Cref{tab:tab2}. We deploy an attention structure with GoogleNet and a classifier for temporal dependency, denoted as the $\mathit{(+)}$\textit{Attention Module}. With the assistance of the weighted values from CNN, there is a 0.008 increment on \datasetfirst\, and at least 0.047 on \datasetsecond, showing the power of CNN as attention. $\mathit{(+)Att_D}$ is the result obtained using softmax along the time and dimension axis to reflect the spatiotemporal importance. The improvement from 0.005 to 0.009 in both datasets indicates that considering the spatial importance is meaningful. The \textit{Key} and \textit{Value Embeddings} strengthen \modelname\, as a linear projection based on \cite{attention}. Although the $\mathit{(+)}$\textit{Positional Encoding} reveals a small performance drop of 0.004 for $\mathit{\tau}$ coefficient and 0.005 for $\mathit{\rho}$ coefficient on \datasetsecond, the performance increases significantly from 0.207 to 0.225 for $\mathit{\tau}$ coefficient and from 0.231 to 0.251 for $\mathit{\rho}$ coefficient on \datasetfirst. $\mathit{(+)X_{CLS}}$ is the result obtained when utilizing the CLS token. Because this study combines the CLS token with adaptive pooling, the CLS token only affects a few video frames. However, adding the CLS token improves the performance on both datasets because it generalizes the classifier, which contains fully connected layers. We also see the effects of skip connection, denoted by $\mathit{(+)}$\textit{Skip Connection}, as suggested by \cite{resnet}. The skip connection exhibits a similar performance on \datasetsecond\, and an improvement of about 0.015 on \datasetfirst.

We also tested different CNN models as the baseline in \Cref{sec:sup_dif_cnn}, various experiments of detailed construction of our model in \Cref{sec:sup_architecture_history}, and several hyperparameters in \Cref{sec:sup_hyperparameter_setting}.

\subsection{Computation Comparison}
\label{exp_computation}

\begin{table}[ht!]
  \centering
  \resizebox{0.475\textwidth}{!}{
    \begin{tabular}{c|ccc|ccc}
      \noalign{\smallskip}\noalign{\smallskip}
      \hline
      \multirow{2}{*}{Method} & \multicolumn{3}{c|}{\datasetfirst} & \multicolumn{3}{c}{\datasetsecond} \\
      % \cline{2-7}
         & Rank & \textit{FE} & \textit{SP} & Rank & \textit{FE} & \textit{SP} \\
      \hline
       DSNet-AF\cite{DSNet}$\mathit{^T}$ & 16 & 413.03G & 1.18G & 13.5 & 661.83G & 1.90G \\
       DSNet-AB\cite{DSNet}$\mathit{^T}$ & 13.5 & 413.03G & 1.29G & 15 & 661.83G & 2.07G \\
       VJMHT\cite{VJMHT}$\mathit{^{T}}$ & 8.5 & 413.03G & 18.21G & 17.5 & 661.83G & 28.25G \\
       VASNet\cite{VASNet}$\mathit{^T}$ & 6 & 413.03G & 1.43G & 9 & 661.83G & 2.30G \\
       RR-STG\cite{RR_STG}$\mathit{^{ST}}$ & 2.5 & 54.82T & 0.31G & 7.5 & 88.41T & 0.20G \\
       MSVA\cite{MSVA}$\mathit{^M}$ & 3.5 & 13.76T & 3.63G & 5.5 & 22.08T & 5.81G \\
       SSPVS\cite{SSPVS}$\mathit{^M}$ & 3 & 413.49G & 20.72G & 4.5 & 662.46G & 44.22G \\
       \bottomrule
       \modelname$\mathit{^{ST}}$ & 1 & 413.03G & 9.78G & 2 & 661.83G & 15.73G
    \end{tabular}
  }
  \caption{Comparison of MACs between video summarization models. Rank is the average rank between Kendall's and Spearman's coefficients in \Cref{tab:tab1}. \textit{FE} is the MACs during feature extraction, and \textit{SP} is that during score predictions. We categorize models as temporal attention-based ($\mathit{T}$), spatiotemporal ($\mathit{ST}$) attention-based, and multi-modal based ($\mathit{M}$) models.}
  \label{tab:tab3}
\end{table}

In this paper, we analyze the computation burdens of video summarization models, focusing on the feature extraction and score prediction steps. The standard procedure for creating summary videos comprises feature extraction, score prediction, and key-shot selection. Feature extraction is a necessary step in converting frames into features using pre-trained models so that video summarization models can take frames of videos as inputs. Score prediction is the step in which video summarization models infer the importance score for videos. Existing studies generally use the same key-shot selection process based on the knapsack algorithm to determine important video segments, so we ignore computations of key-shot selection.

\Cref{tab:tab3} displays MACs measurements and compares the computation resources during the inference per video. \modelname\, performs best with relatively fewer MACs than the other video summarization models. Based on the average rank from \Cref{tab:tab1}, more computational costs or supplemental data from other modalities is inevitable for better video summarization performance. Unlike previous approaches, \modelname\, exhibits high performance with fewer computational resources by exploiting CNN as a sliding window.

We find that our model is more efficient than previous ones when considering spatiotemporal contexts. RR-STG \cite{RR_STG} shows much fewer MACs than \modelname\, during score predictions; however, it shows exceptionally more MACs during feature extraction than others. RR-STG utilizes feature extraction steps for visual relationships by inputting each frame into the object detection model \cite{faster_rcnn}, thereby, relying heavily on the pre-processing steps. While summarizing the new videos, RR-STG needs significant time to get spatial associations even though the score prediction takes less time.
Other methods \cite{DMASum, STVT, DAN, VSS_Net} design two modules to reflect spatial and temporal dependency, respectively, as shown in \Cref{fig:figure1-a} and \Cref{fig:figure1-b}. These approaches become costly when processing numerous frames in long videos for video summarization. \modelname\, effectively captures spatiotemporal importance in one way using CNN, as illustrated in \Cref{fig:figure1-c}. Thus, our proposed method shows superior performance by focusing on temporal and visual importance.
\section{Conclusion}
\raggedbottom
\label{sec:conclusion}

This study addresses the problem of attention in video summarization. The existing pairwise attention-based video summarization mechanisms fail to account for visual dependencies, and prior research addressing this issue involves significant computational demands. To deal with the same problem efficiently, we propose \modelname, in which a CNN’s ability is used for video summarization for the first time. We also verify that the CNN works on frame features and creates attention maps. The strength of the CNN allows \modelname\, to achieve state-of-the-art results based on the overall performance of two popular benchmark datasets with fewer MACs than before. Our proposed model even outperforms multi-modal or external dataset-based models without additional data. For future work, we suggest further exploring how CNN affects video representations by tailoring frame feature-specific CNN models or training feature-extraction and attention-based CNN models. We believe this study can encourage follow-up research on video summarization and other video-related deep-learning studies.

\paragraph{Acknowledgements.} This work was supported by Korea Internet \& Security Agency(KISA) grant funded by the Korea government(PIPC) (No.RS-2023-00231200, Development of personal video information privacy protection technology capable of AI learning in an autonomous driving environment)
{
    \small
    \bibliographystyle{ieeenat_fullname}
    \bibliography{main}
}
\clearpage
\setcounter{page}{1}
\appendix
\maketitlesupplementary

\section{Experiment details}
\label{sec:sup_experiment_details}
\subsection{Measure correlation}
\label{sec:sup_measure_correlation}
Based on \cite{MAAM}, we test everything 10 times in each experiment for strict evaluation since video summarization models are sensitive to randomness due to the lack of datasets. Additionally, we follow \cite{MAAM} to perform the experiments rigorously by using non-overlapping five-fold cross-validation for reflection of all videos as test data. For each fold, we use 80\% of the videos in the dataset for training and 20\% for testing. We then average the results of all folds to export the final score. Owing to non-overlapping videos in the training data in each split, different training epochs are required; therefore, we pick the model that shows the best performance on test data during the training epochs of each split. During training, the predicted score for each input video is compared to the average score of all ground truth scores of summary videos for that input video. During inference, the performance for each video is calculated by comparing each ground truth score with the predicted score and then averaging them.

\subsection{Implementation details}
\label{sec:sup_implementation_details}
For a fair comparison, we follow the standard procedure \cite{vsLSTM, SUM_GAN, STVT, MSVA, SSPVS, VJMHT} by uniformly subsampling the videos to 2 fps and acquiring the image representation of every frame from GoogleNet \cite{googlenet}. GoogleNet is also used as a trainable CNN to match the dimension of all features to 1,024, and all CNN models are pre-trained on ImageNet \cite{imagenet}. The initial weights of the linear layers in the classifier are initialized by Xavier initialization \cite{xavier_initialization}, while key and value embeddings are initialized randomly. The output channels of linear layers and key and value embedding dimensions are 1,024. The reduction ratio $\mathit{r}$ in CNN is 32, an inherent trait of GoogleNet, and all adaptive pooling layers are adaptive average pooling operations. The shape of the CLS token is 3$\times$1,024, the epsilon value for layer normalization is 1e-6, and the dropout rate is 0.6. We train \modelname\, on a single NVIDIA GeForce RTX 4090 for 100 epochs with a batch size of 1 and use an Adam optimizer \cite{adam} with 1e-3 as the learning rate and 1e-7 as weight decay.

\section{Summary video visualization}
\label{sec:sup_visualize_sum}

\begin{figure*}[ht!]% >>>
  \centering
  \begin{subfigure}{\textwidth}
    \centering
    \captionsetup{justification=centering}
    \includegraphics[width=\textwidth]{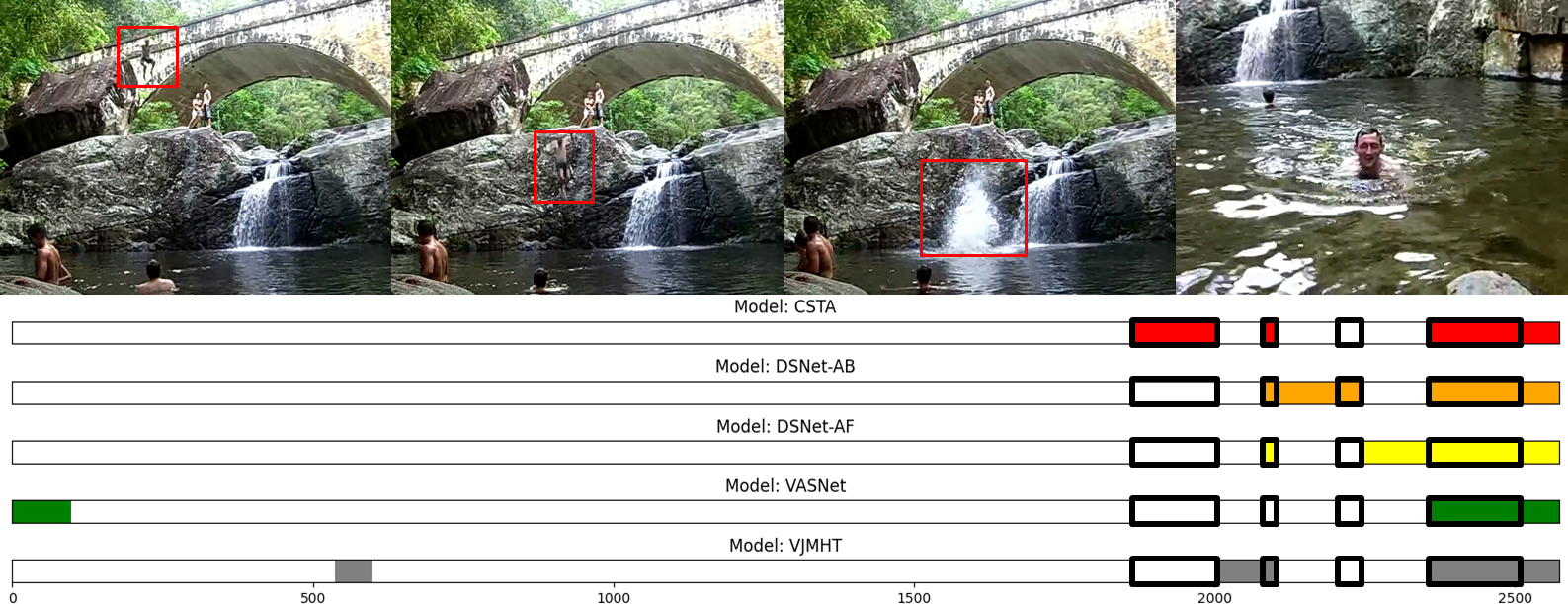}
    \caption{The images from the summary video titled ``paluma jump" about people diving into the water.}
    \label{fig:figure5-a}
  \end{subfigure}
\end{figure*}

\begin{figure*}\ContinuedFloat
  \begin{subfigure}{\textwidth}
    \centering
    \captionsetup{justification=centering}
    \includegraphics[width=\textwidth]{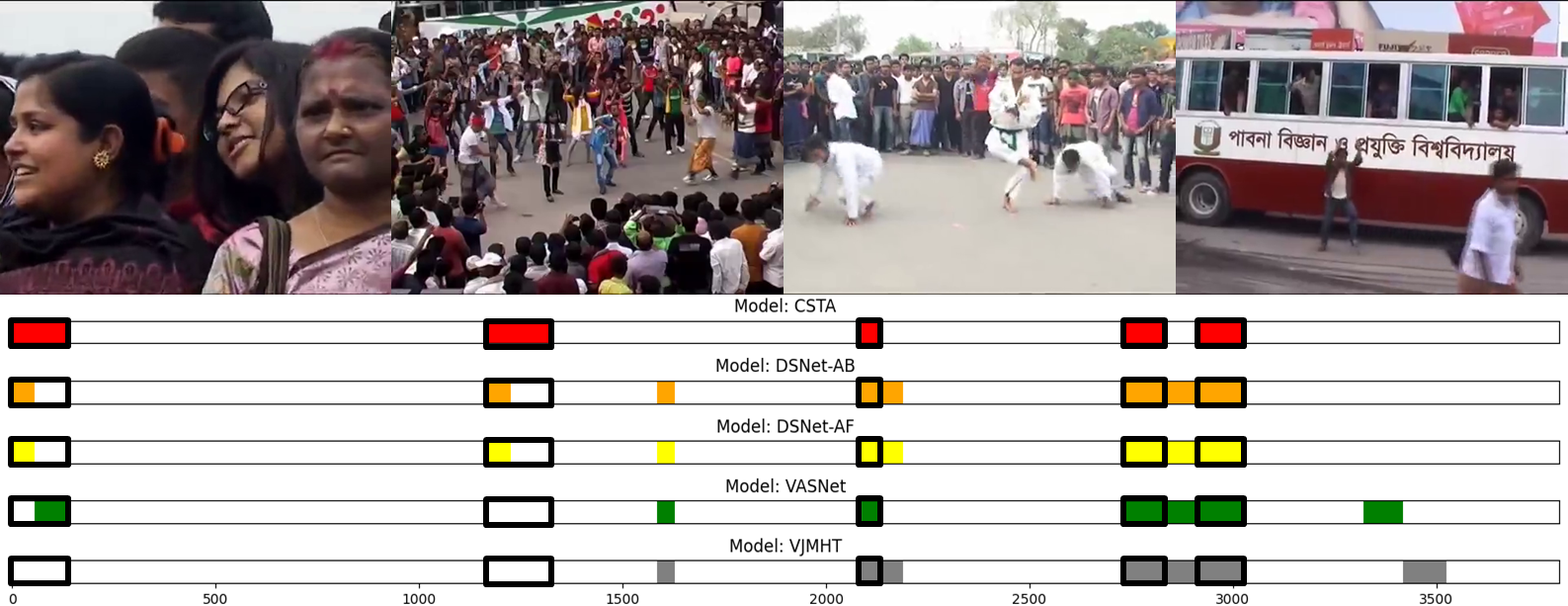}
    \caption{The images from the summary video titled ``ICC World Twenty20 Bangladesh 2014 Flash Mob - Pabna University of Science \& Technology ( PUST )" about people performing flash mobs on the street and crowds watching them.}
    \label{fig:figure5-b}
  \end{subfigure}
  \\[\baselineskip]
  \begin{subfigure}{\textwidth}
    \centering
    \captionsetup{justification=centering}
    \includegraphics[width=\textwidth]{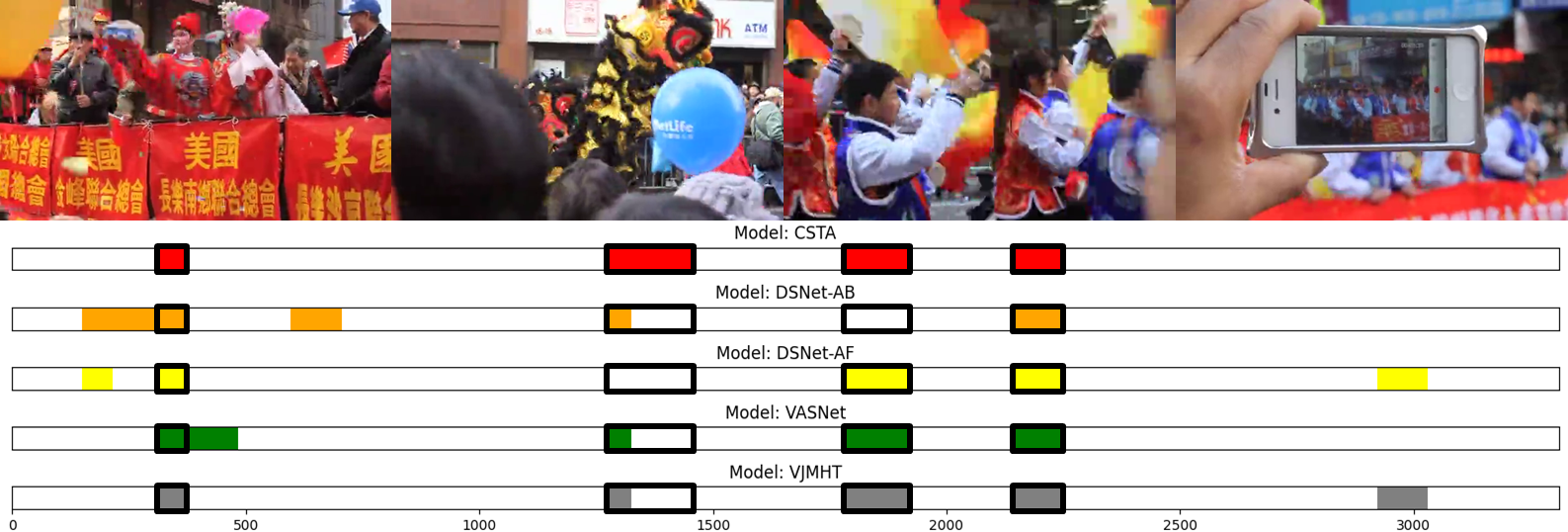}
    \caption{The images from the summary video titled ``Chinese New Year Parade 2012 New York City Chinatown" about the parade celebrating the Chinese New Year on the streets of New York City.}
    \label{fig:figure5-c}
  \end{subfigure}
  \caption
    {%
      Visualization and comparison of summary videos generated by different models. The images above are the frames selected by \modelname\, as parts of the summary video. The graphs below show which frames models pick as keyframes. From the graphs, each row is the result of each model. The x-axis is the order of the frames, and the black boxes are the ground truth frames. The color parts are the frames each model selects, and the white parts are the frames unselected by each model.
    }%
  \label{fig:figure5}
\end{figure*}%

As shown in \Cref{fig:figure5}, we visualize and compare the generated summary videos from different models. We compared \modelname\, with DSNet-AB, DSNet-AF \cite{DSNet}, VASNet \cite{VASNet}, and VJMHT \cite{VJMHT}. The videos were selected from \datasetfirst\, (\Cref{fig:figure5-a}) and \datasetsecond\, (\Cref{fig:figure5-b} and \Cref{fig:figure5-c}). Since each model used different videos for the test, we chose videos used for training by all models.\par

The summary video in \Cref{fig:figure5-a} was taken during the ``paluma jump," representing people diving into the water at Paluma. The first three frames show the exact moment people dive into the water. Based on the selected frames in the graphs, \modelname\, selects keyframes that represent the main content of the video more accurately than the other models. Although other models chose key moments in the later part of the video, they did not search for diving moments as precisely as \modelname.

The summary video in \Cref{fig:figure5-b} was taken during the ``ICC World Twenty20 Bangladesh 2014 Flash Mob - Pabna University of Science \& Technology ( PUST )," representing flash mops on the street. The frames selected by \modelname\, display different flash mop performances on the street and the people watching them. \modelname\, selects keyframes in videos more often than the other models, which either select non-keyframes or skip keyframes, as shown in the graphs in \Cref{fig:figure5-b}.

The summary video in \Cref{fig:figure5-c} was taken during the ``Chinese New Year Parade 2012 New York City Chinatown," representing the parade celebrating the Chinese New Year in New York City. Based on the chosen images, \modelname\, finds representative frames containing the parade or people reacting to it (\eg, images showing tiger-like masks, people marching on the street, or people recording the parade, respectively). Unlike the other models, the graphs exhibit \modelname\, creating exactly the same summary videos with ground truth. These results suggest the superiority of \modelname\, over the other models.

%%%%%%%%%%%%%%%%%%%%%%%%%%%%%%%%%%%%%%%%%%%%%%%%%%%%%%%%%%%%%%%%%%%%%%%%%%%%%%%%%%%%%%%%%%%%%%%%%
%%%%%%%%%%%%%%%%%%%%%%%%%%%%%%%%%%%%%%%%%%%%%%%%%%%%%%%%%%%%%%%%%%%%%%%%%%%%%%%%%%%%%%%%%%%%%%%%%
%%%%%%%%%%%%%%%%%%%%%%%%%%%%%%%%%%%%%%%%%%%%%%%%%%%%%%%%%%%%%%%%%%%%%%%%%%%%%%%%%%%%%%%%%%%%%%%%%

\section{CNN models}
\label{sec:sup_dif_cnn}

\begin{table}[ht]
  \centering
  \resizebox{0.9\textwidth}{!}{
    \begin{tabular}{l|cc|cc|l|cc|cc}
      \noalign{\smallskip}\noalign{\smallskip}
      \hline
      \multirow{2}{*}{Baseline} & \multicolumn{2}{c|}{\datasetfirst} & \multicolumn{2}{c|}{\datasetsecond} & \multirow{2}{*}{\modelname} & \multicolumn{2}{c|}{\datasetfirst} & \multicolumn{2}{c}{\datasetsecond} \\
      & $\mathit{\tau}$ & $\mathit{\rho}$ & $\mathit{\tau}$ & $\mathit{\rho}$ & & $\mathit{\tau}$ & $\mathit{\rho}$ & $\mathit{\tau}$ & $\mathit{\rho}$ \\
      \hline
       MobileNet-V2\cite{mobilenet} & 0.170 & 0.189 & 0.122 & 0.155 & \modelname(MobileNet-V2) & 0.228 & 0.255 & 0.194 & 0.254 \\
       EfficientNet-B0\cite{efficientnet} & 0.185 & 0.206 & 0.119 & 0.150 & \modelname(EfficientNet-B0) & 0.222 & 0.247 & 0.194 & 0.255 \\
       ResNet-18\cite{resnet} & 0.167 & 0.187 & 0.140 & 0.178 & \modelname(ResNet-18) & 0.225 & 0.251 & 0.195 & 0.256 \\
    \end{tabular}
  }
  \caption{The results of \modelname\, with different CNN models as the baseline.}
  \label{tab:tab4}
\end{table}

We tested \modelname\, using different CNN models as the baseline, as shown in \Cref{tab:tab4}. We unified the dimension size to 1,024 because each CNN model exports different dimensions of features. All CNN models improved their performance with the \modelname\, architecture. This supports the notion that \modelname\, does not work only for GoogleNet.

\section{Architecture history}
\label{sec:sup_architecture_history}
Here, we provide a step-by-step explanation of how \modelname\, is constructed. For all experiments, $\mathit{\tau}$ and $\mathit{\rho}$ represent Kendall's and Spearman's coefficients, respectively. The score marked in bold indicates that the model was selected because it yielded the best performance from the experiment. Each experiment was tested 10 times for strict verification, and the average score was recorded as the final one.\par

As explained in \Cref{sec:exp_setting}, the form of the ground truth of \datasetfirst\, is summary videos, so the models aim to generate summary videos correctly. Kendall's and Spearman's coefficients between the predicted and ground truth summary videos are the basis for evaluating the performance of \datasetfirst. Based on the code provided in previous studies \cite{A2Summ, MSVA}, we generate summary videos by assigning 1 for selected frames and 0 otherwise. In videos, most frames are not keyframes, so the performance of \datasetfirst\, is usually higher than that of \datasetsecond. The form of the ground truth of \datasetsecond\, is the shot-level importance score, so models should aim to predict accurate shot-level importance scores. The scores of entire frames are determined by assigning the identical scores of subsampled frames to nearby frames based on the code provided in previous studies \cite{A2Summ, DSNet, pglsum, VASNet}. Therefore, Kendall's and Spearman's coefficients of subsampled frames are the basis for evaluating the performance of \datasetsecond. The difference between \datasetfirst\, and \datasetsecond\, can cause bad performance on \datasetsecond, and the performance on \datasetfirst\, looks much better than that on \datasetsecond.

\subsection{Channel of input feature}
\label{sec:sup_input_channel}

\begin{table}[ht]
  \centering
  \resizebox{0.4\textwidth}{!}{
    \begin{tabular}{l|cc|cc}
      \noalign{\smallskip}\noalign{\smallskip}
      \hline
      \multirow{2}{*}{Channels} & \multicolumn{2}{c|}{SumMe} & \multicolumn{2}{c}{TVSum} \\
         & $\mathit{\tau}$ & $\mathit{\rho}$ & $\mathit{\tau}$ & $\mathit{\rho}$ \\
      \hline
       1 & 0.169 & 0.188 & 0.122 & 0.154 \\
       3 & 0.176 & 0.197 & 0.129 & 0.163 \\
    \end{tabular}
  }
  \caption{Comparison of different number of input channels on GoogleNet.}
  \label{tab:tab5}
\end{table}

In \Cref{tab:tab5}, we test the number of channels of input frame features. As explained in \Cref{sec:method_embedding}, we copy the input frame feature two times to create three channels of input to match the number of common channels of images that are usually used to train CNN models. We use GoogleNet as the baseline and check the results when the channels of input frame features are 1 and 3. The model taking 3 channels of features as inputs performs better than taking a single channel of features. This supports the idea that creating the shape of input frame features the same as the RGB images helps to utilize CNN models better.

\subsection{Verify attention}
\label{sec:sup_verify_att}

\paragraph{Softmax.}

\begin{table}[ht]
  \centering
  \resizebox{0.5\textwidth}{!}{
    \begin{tabular}{l|cc|cc}
      \noalign{\smallskip}\noalign{\smallskip}
      \hline
      \multirow{2}{*}{Setting} & \multicolumn{2}{c|}{\datasetfirst} & \multicolumn{2}{c}{\datasetsecond} \\
      % \cline{2-7}
         & $\mathit{\tau}$ & $\mathit{\rho}$ & $\mathit{\tau}$ & $\mathit{\rho}$ \\
      \hline
       Baseline(GoogleNet) & 0.176 & 0.197 & 0.129 & 0.163 \\
       \textbf{Baseline$\mathit{\textbf{+Att}}$} & \textbf{0.214} & \textbf{0.239} & \textbf{0.167} & \textbf{0.219} \\
       Baseline$\mathit{+Att+Soft_{T}}$ & 0.184 & 0.205 & 0.176 & 0.231 \\
       Baseline$\mathit{+Att+Soft_{D}}$ & 0.186 & 0.207 & 0.170 & 0.224 \\
       \textbf{Baseline$\mathit{\textbf{+Att+Soft}_{\textbf{T\&D}}}$} & \textbf{0.189} & \textbf{0.211} & \textbf{0.182}& \textbf{0.240} \\
    \end{tabular}
  }
  \caption{The ablation study for the softmax. The baseline is the plain GoogleNet summarizing videos, which is the same as in \Cref{fig:figure4}. $\mathit{Att}$ is an attention-based CNN structure without softmax. $\mathit{Soft}$ applies the softmax operation to the model along the frame axis ($\mathit{T}$), dimension axis ($\mathit{D}$), or both axes ($\mathit{T\&D}$).}
  \label{tab:tab6}
\end{table}

We verify the effects of attention structure and perform softmax along different axes, as shown in \Cref{tab:tab6}. Attention structure, composed of CNN generating attention maps and the classifier predicting scores, increases the baseline performance regardless of softmax (Baseline$\mathit{+Att}$). Compared to the baseline, the attention structure increased by 0.038 and 0.042 for Kendall's and Spearman's coefficients, respectively, on \datasetfirst, where it increased by 0.038 and 0.056 for Kendall's and Spearman's coefficients, respectively, on \datasetsecond. This demonstrates CNN's ability as an attention algorithm.\par

Reflecting weighted values between frames ($\mathit{Att+Soft_{T}}$) or dimensions ($\mathit{Att+Soft_{D}}$) improved the baseline model by at least 0.008 on \datasetfirst\, and 0.041 on \datasetsecond. This supports the importance of spatial attention, and it is better to consider weighted values along both the frame and dimension axes ($\mathit{Att+Soft_{T\&D}}$) than focusing on only one of the axes.\par

On \datasetfirst, the model without softmax is better than the model with softmax; however, the reverse is the case on \datasetsecond. Since there is no best model for all datasets, we choose both models as the baseline and find the best one when extending the structure.

\paragraph{Balance Ratio.}

\begin{table}[ht]
  \centering
  \resizebox{0.5\textwidth}{!}{
    \begin{tabular}{l|cc|cc}
      \noalign{\smallskip}\noalign{\smallskip}
      \hline
      \multirow{2}{*}{Setting} & \multicolumn{2}{c|}{\datasetfirst} & \multicolumn{2}{c}{\datasetsecond} \\
      % \cline{2-7}
         & $\mathit{\tau}$ & $\mathit{\rho}$ & $\mathit{\tau}$ & $\mathit{\rho}$ \\
      \hline
       \textbf{Baseline}($\mathit{\textbf{Att}_{\textbf{T\&D}}}$) & \textbf{0.189} & \textbf{0.211} & \textbf{0.182} & \textbf{0.240} \\
       Baseline$\mathit{+Balance_T}$ & 0.186 & 0.207 & 0.178 & 0.233 \\
       Baseline$\mathit{+Balance_D}$ & 0.186 & 0.207 & 0.182 & 0.239 \\
       Baseline$\mathit{+Balance_{BD}}$ & 0.186 & 0.207 & 0.175 & 0.230 \\
       Baseline$\mathit{+Balance_{BU}}$ & 0.187 & 0.208 & 0.183 & 0.240 \\
    \end{tabular}
  }
  \caption{The ablation study for balancing ratio. The baseline applies the softmax operation to the attention map along the frame and dimension axes (\Cref{tab:tab6}). $\mathit{Balance}$ is the balancing ratio between frames and dimensions. $\mathit{T}$ adjusts the weighted values along the frame axis to the dimension axis. $\mathit{D}$ adjusts the scale of the weighted values along the dimension axis to the frame axis. $\mathit{BD}$ decreases the scale of larger ones into smaller ones. $\mathit{BU}$ upscales the scale of smaller ones into larger ones.}
  \label{tab:tab7}
\end{table}

We hypothesize that the imbalance ratio between frames and dimensions deteriorates the performance of the model with softmax. For example, suppose the number of frames is 100, and the dimension size is 1,000. In this case, the weighted values between frames are usually larger than those between dimensions (on average, 0.01 and 0.001 between frames and dimensions, respectively). This situation can lead to overfitting the frame importance, so we tested the performance of the model under a balanced ratio between the number of frames and dimensions, as shown in \Cref{tab:tab7}. However, all results were worse than the baseline, so we used the default setting.\par

%%%%%%%%%%%%%%%%%%%%%%%%%%%%%%%%%%%%%%%%%%%%%%%%%%%%%%%%%%%%%%%%%%%%%%%%%%%%%%%%%%%%%%%%%%%%%%%%%
%%%%%%%%%%%%%%%%%%%%%%%%%%%%%%%%%%%%%%%%%%%%%%%%%%%%%%%%%%%%%%%%%%%%%%%%%%%%%%%%%%%%%%%%%%%%%%%%%
%%%%%%%%%%%%%%%%%%%%%%%%%%%%%%%%%%%%%%%%%%%%%%%%%%%%%%%%%%%%%%%%%%%%%%%%%%%%%%%%%%%%%%%%%%%%%%%%%

\subsection{Self-attention extension}
\label{sec:sup_sa_ext}

\begin{table}[ht]
  \centering
  \begin{subtable}[c]{\textwidth}
    \centering
    \resizebox{0.99\textwidth}{!}{
      \begin{tabular}{l|cc|cc|l|cc|cc}
        \noalign{\smallskip}\noalign{\smallskip}
        \hline
        \multirow{2}{*}{Setting} & \multicolumn{2}{c|}{\datasetfirst} & \multicolumn{2}{c|}{\datasetsecond} & \multirow{2}{*}{Setting} & \multicolumn{2}{c|}{\datasetfirst} & \multicolumn{2}{c}{\datasetsecond} \\
        % \cline{2-7}
        & $\mathit{\tau}$ & $\mathit{\rho}$ & $\mathit{\tau}$ & $\mathit{\rho}$ & & $\mathit{\tau}$ & $\mathit{\rho}$ & $\mathit{\tau}$ & $\mathit{\rho}$ \\
        \hline
         Baseline($\mathit{Att}$) & 0.214 & 0.239 & 0.167 & 0.219 & Baseline$\mathit{+EMB}$ & 0.158 & 0.177 & 0.033 & 0.042 \\
         Baseline$\mathit{+Scale_{D}}$ & 0.220 & 0.246 & 0.173 & 0.227 & Baseline$\mathit{+EMB+Scale_{D}}$ & 0.209 & 0.238 & 0.187 & 0.246 \\
         Baseline$\mathit{+Scale_{T}}$ & 0.213 & 0.238 & 0.173 & 0.227 & \textbf{Baseline}$\mathit{\textbf{+EMB+Scale}_{\textbf{T}}}$ & \textbf{0.214} & \textbf{0.239} & \textbf{0.187} & \textbf{0.245} \\
         Baseline$\mathit{+Scale_{T\&D}}$ & 0.196 & 0.218 & 0.154 & 0.203 & Baseline$\mathit{+EMB+Scale_{T\&D}}$ & 0.192 & 0.215 & 0.191 & 0.250 \\
      \end{tabular}
    }
    \caption{The result of the model without softmax as the baseline.}
    \label{tab:tab8-a}
  \end{subtable}
  \begin{subtable}[c]{\textwidth}
    \centering
    \resizebox{0.99\textwidth}{!}{
      \begin{tabular}{l|cc|cc|l|cc|cc}
        \noalign{\smallskip}\noalign{\smallskip}
        \hline
        \multirow{2}{*}{Setting} & \multicolumn{2}{c|}{\datasetfirst} & \multicolumn{2}{c|}{\datasetsecond} & \multirow{2}{*}{Setting} & \multicolumn{2}{c|}{\datasetfirst} & \multicolumn{2}{c}{\datasetsecond} \\
        % \cline{2-7}
        & $\mathit{\tau}$ & $\mathit{\rho}$ & $\mathit{\tau}$ & $\mathit{\rho}$ & & $\mathit{\tau}$ & $\mathit{\rho}$ & $\mathit{\tau}$ & $\mathit{\rho}$ \\
        \hline
         Baseline($\mathit{Att_{T\&D}}$) & 0.189 & 0.211 & 0.182 & 0.240 & \textbf{Baseline}$\mathit{\textbf{+EMB}}$ & \textbf{0.207} & \textbf{0.231} & \textbf{0.193} & \textbf{0.253} \\
         Baseline$\mathit{+Scale_{D}}$ & 0.151 & 0.168 & 0.192 & 0.251 & Baseline$\mathit{+EMB+Scale_{D}}$ & 0.162 & 0.181 & 0.190 & 0.249 \\
         Baseline$\mathit{+Scale_{T}}$ & 0.160 & 0.178 & 0.192 & 0.252 & Baseline$\mathit{+EMB+Scale_{T}}$ & 0.163 & 0.182 & 0.191 & 0.251 \\
         Baseline$\mathit{+Scale_{T\&D}}$ & 0.149 & 0.166 & 0.187 & 0.146 & Baseline$\mathit{+EMB+Scale_{T\&D}}$ & 0.163 & 0.181 & 0.187 & 0.244 \\
      \end{tabular}
    }
    \caption{The result of the model with softmax as the baseline.}
    \label{tab:tab8-b}
  \end{subtable}
  \caption{The ablation study for methods in self-attention. $\mathit{Att}$ is the model without softmax, and $\mathit{Att_{T\&D}}$ is the model with softmax along the frame and dimension axes (\Cref{tab:tab6}). $\mathit{EMB}$ employs key and value embedding into the baseline model. $\mathit{Scale}$ divides the values of attention maps by the number of frames ($\mathit{T}$) or dimensions ($\mathit{D}$) or both of them ($\mathit{T\&D}$).}
  \label{tab:tab8}
\end{table}

Given that our model operates the attention structure differently from existing ones, we must test which existing method works for \modelname. First, we verify the key, value embeddings, and scaling factors used in self-attention \cite{attention}. The key and value embeddings project input data into another space by exploiting linear layers. At the same time, the scaling factor divides all values of attention maps with the size of the dimension. Unlike self-attention handling 1-dimensional data, we should consider the frame and dimension axes for the scaling factor because of 2-dimensional data. We test the scaling factor using the size of dimensions ($\mathit{Scale_{D}}$), frames ($\mathit{Scale_{T}}$), and both ($\mathit{Scale_{T\&D}}$).\par

The best performance for the model without softmax is achieved by utilizing the key, value embedding, and scaling factors with the size of frames ($\mathit{EMB+Scale_{T}}$), as shown in \Cref{tab:tab8-a}. Although utilizing the scaling factors with the size of dimensions ($\mathit{Scale_{D}}$) yields better performance than $\mathit{EMB+Scale_{T}}$ on \datasetfirst, it yields much worse performance on \datasetsecond, even considering the performance gaps on \datasetfirst. Also, $\mathit{EMB+Scale_{D}}$ and $\mathit{EMB+Scale_{T\&D}}$ show slightly better performance than $\mathit{EMB+Scale_{T}}$ on \datasetsecond, but much worse on \datasetfirst. We select $\mathit{EMB+Scale_{T}}$ as the best model based on overall performance.\par

For the model with softmax, we select the model using key and value embedding (Baseline$\mathit{+EMB}$) because it reveals the best performance for all datasets, as shown in \Cref{tab:tab8-b}.

\subsection{Transformer extension}
\label{sec:sup_tf_ext}

\begin{table}[ht]
  \centering
  \begin{subtable}[c]{\textwidth}
    \centering
    \resizebox{0.99\textwidth}{!}{
      \begin{tabular}{l|cc|cc|l|cc|cc}
        \noalign{\smallskip}\noalign{\smallskip}
        \hline
        \multirow{2}{*}{Setting} & \multicolumn{2}{c|}{\datasetfirst} & \multicolumn{2}{c|}{\datasetsecond} & \multirow{2}{*}{Setting} & \multicolumn{2}{c|}{\datasetfirst} & \multicolumn{2}{c}{\datasetsecond} \\
        % \cline{2-7}
         & $\mathit{\tau}$ & $\mathit{\rho}$ & $\mathit{\tau}$ & $\mathit{\rho}$ & & $\mathit{\tau}$ & $\mathit{\rho}$ & $\mathit{\tau}$ & $\mathit{\rho}$ \\
        \hline
         \textbf{Baseline}($\mathit{\textbf{Att}}$) & \textbf{0.214} & \textbf{0.239} & \textbf{0.187} & \textbf{0.245} & Baseline$\mathit{+Drop}$ & 0.181 & 0.202 & 0.190 & 0.250 \\
         Baseline$\mathit{+FPE(TD)}$ & 0.145 & 0.161 & 0.031 & 0.041 & Baseline$\mathit{+FPE(TD)+Drop}$ & 0.148 & 0.165 & -0.007 & -0.009 \\
         Baseline$\mathit{+FPE(T)}$ & 0.192 & 0.214 & 0.189 & 0.248 & Baseline$\mathit{+FPE(T)+Drop}$ & 0.179 & 0.200 & 0.191 & 0.251 \\
         Baseline$\mathit{+RPE(TD)}$ & 0.193 & 0.216 & 0.190 & 0.249 & Baseline$\mathit{+RPE(TD)+Drop}$ & 0.172 & 0.192 & 0.191 & 0.250 \\
         Baseline$\mathit{+RPE(T)}$ & 0.195 & 0.217 & 0.191 & 0.251 & Baseline$\mathit{+RPE(T)+Drop}$ & 0.190 & 0.211 & 0.188 & 0.247 \\
         Baseline$\mathit{+LPE(TD)}$ & 0.140 & 0.156 & -0.032 & -0.043 & Baseline$\mathit{+LPE(TD)+Drop}$ & 0.136 & 0.151 & -0.004 & -0.005 \\
         Baseline$\mathit{+LPE(T)}$ & 0.152 & 0.169 & 0.062 & 0.082 & Baseline$\mathit{+LPE(T)+Drop}$ & 0.138 & 0.153 & 0.036 & 0.048 \\
         Baseline$\mathit{+CPE(TD)}$ & 0.143 & 0.160 & 0.168 & 0.221 & Baseline$\mathit{+CPE(TD)+Drop}$ & 0.139 & 0.155 & 0.175 & 0.229 \\
         Baseline$\mathit{+CPE(T)}$ & 0.142 & 0.158 & 0.140 & 0.183 & Baseline$\mathit{+CPE(T)+Drop}$ & 0.140 & 0.156 & 0.145 & 0.190 \\
      \end{tabular}
    }
    \caption{The results of the model without softmax as the baseline. $\mathit{Att}$ applies key and value embedding with scaling factor for the size of frames without softmax (\Cref{tab:tab8-a}).}
    \label{tab:tab9-a}
  \end{subtable}
  \begin{subtable}[c]{\textwidth}
    \centering
    \resizebox{0.99\textwidth}{!}{
      \begin{tabular}{l|cc|cc|l|cc|cc}
        \noalign{\smallskip}\noalign{\smallskip}
        \hline
        \multirow{2}{*}{Setting} & \multicolumn{2}{c|}{\datasetfirst} & \multicolumn{2}{c|}{\datasetsecond} & \multirow{2}{*}{Setting} & \multicolumn{2}{c|}{\datasetfirst} & \multicolumn{2}{c}{\datasetsecond} \\
         & $\mathit{\tau}$ & $\mathit{\rho}$ & $\mathit{\tau}$ & $\mathit{\rho}$ & & $\mathit{\tau}$ & $\mathit{\rho}$ & $\mathit{\tau}$ & $\mathit{\rho}$ \\
        \hline
         \textbf{Baseline}($\mathit{\textbf{Att}_{\textbf{T\&D}}}$) & \textbf{0.207} & \textbf{0.231} & \textbf{0.193} & \textbf{0.253} & Baseline$\mathit{+Drop}$ & 0.134 & 0.149 & 0.192 & 0.252 \\
         Baseline$\mathit{+FPE(TD)}$ & 0.144 & 0.160 & 0.152 & 0.199 & Baseline$\mathit{+FPE(TD)+Drop}$ & 0.148 & 0.165 & 0.134 & 0.176 \\
         Baseline$\mathit{+FPE(T)}$ & 0.138 & 0.154 & 0.191 & 0.250 & Baseline$\mathit{+FPE(T)+Drop}$ & 0.152 & 0.170 & 0.193 & 0.253 \\
         Baseline$\mathit{+RPE(TD)}$ & 0.149 & 0.166 & 0.193 & 0.253 & Baseline$\mathit{+RPE(TD)+Drop}$ & 0.151 & 0.168 & 0.193 & 0.253 \\
         Baseline$\mathit{+RPE(T)}$ & 0.142 & 0.159 & 0.192 & 0.252 & Baseline$\mathit{+RPE(T)+Drop}$ & 0.158 & 0.176 & 0.192 & 0.252 \\
         Baseline$\mathit{+LPE(TD)}$ & 0.152 & 0.169 & 0.078 & 0.102 & Baseline$\mathit{+LPE(TD)+Drop}$ & 0.148 & 0.164 & 0.052 & 0.068 \\
         Baseline$\mathit{+LPE(T)}$ & 0.134 & 0.149 & 0.079 & 0.103 & Baseline$\mathit{+LPE(T)+Drop}$ & 0.145 & 0.162 & 0.089 & 0.117 \\
         Baseline$\mathit{+CPE(TD)}$ & 0.135 & 0.150 & 0.119 & 0.156 & Baseline$\mathit{+CPE(TD)+Drop}$ & 0.146 & 0.162 & 0.114 & 0.149 \\
         Baseline$\mathit{+CPE(T)}$ & 0.143 & 0.159 & 0.180 & 0.236 & Baseline$\mathit{+CPE(T)+Drop}$ & 0.149 & 0.166 & 0.184 & 0.241 \\
      \end{tabular}
    }
    \caption{The results of the model with softmax as the baseline. $\mathit{Att_{T\&D}}$ applies key and value embedding with softmax (\Cref{tab:tab8-b}).}
    \label{tab:tab9-b}
  \end{subtable}
  \caption{The ablation study for methods in transformers. $\mathit{FPE}$ is fixed positional encoding, $\mathit{RPE}$ is relative positional encoding, $\mathit{LPE}$ is learnable positional encoding, and $\mathit{CPE}$ is conditional positional encoding. $\mathit{T}$ represents the frame axis, and $\mathit{TD}$ represents the frame and dimension axis for positional encoding. $\mathit{Drop}$ exploits dropout to input frame features.}
  \label{tab:tab9}
\end{table}

We verify the methods used in transformers \cite{attention}, which are positional encodings and dropouts. Positional encoding strengthens position awareness, whereas dropout enhances generalization. We expect the same effects when we apply the positional encodings and dropouts to the input frame features. We use fixed positional encoding ($\mathit{FPE}$) \cite{attention}, relative positional encoding ($\mathit{RPE}$) \cite{pglsum}, learnable positional encoding ($\mathit{LPE}$) \cite{attention}, and conditional positional encoding ($\mathit{CPE}$) \cite{CPVT}. We must test both 2-dimensional ($\mathit{TD}$) and 1-dimensional ($\mathit{T}$) positional encoding matrices to represent temporal position explicitly because the data structure differs from the original positional encoding that handles only 1-dimensional data. For $\mathit{CPE}$, $\mathit{T}$ operates a depth-wise 1D CNN operation for each channel, whereas $\mathit{TD}$ operates entire channels. We use 0.1 as the dropout ratio, the same as \cite{attention}. \par

The results of both models with and without softmax reveal that employing positional encodings or dropout into the input frame features deteriorates the performance of Kendall's and Spearman's coefficients for all datasets, as shown in \Cref{tab:tab9}. We suppose that adding different values to each frame feature leads to distortion of data, making it difficult for the model to learn patterns from frame features because \modelname\, considers the frame features as images. If more data are available, the model can learn location information from these positional encodings because they are similar to bias. Thus, all results yielded by the model with and without softmax worsen when using positional encoding or dropout on \datasetfirst. However, some results are similar to or even better than the baseline on \datasetsecond\, because \datasetsecond\, has more data than \datasetfirst. Due to the lack of data, we chose the baseline models based on performance.

\subsection{PGL-SUM extension}
\label{sec:sup_pgl_ext}

\begin{table}[ht]
  \centering
  \begin{subtable}[c]{\textwidth}
    \centering
    \resizebox{0.99\textwidth}{!}{
      \begin{tabular}{l|cc|cc|l|cc|cc}
        \noalign{\smallskip}\noalign{\smallskip}
        \hline
        \multirow{2}{*}{Setting} & \multicolumn{2}{c|}{\datasetfirst} & \multicolumn{2}{c|}{\datasetsecond} & \multirow{2}{*}{Setting} & \multicolumn{2}{c|}{\datasetfirst} & \multicolumn{2}{c}{\datasetsecond} \\
          & $\mathit{\tau}$ & $\mathit{\rho}$ & $\mathit{\tau}$ & $\mathit{\rho}$ & & $\mathit{\tau}$ & $\mathit{\rho}$ & $\mathit{\tau}$ & $\mathit{\rho}$ \\
        \hline
         Baseline($\mathit{Att}$) & 0.214 & 0.239 & 0.187 & 0.245 & Baseline$\mathit{+Drop}$ & 0.219 & 0.244 & 0.191 & 0.250 \\
         Baseline$\mathit{+FPE(TD)}$ & 0.118 & 0.131 & 0.064 & 0.084 & Baseline$\mathit{+FPE(TD)+Drop}$ & 0.164 & 0.183 & 0.149 & 0.197 \\
         Baseline$\mathit{+FPE(T1)}$ & 0.184 & 0.205 & 0.169 & 0.222 & Baseline$\mathit{+FPE(T1)+Drop}$ & 0.183 & 0.204 & 0.171 & 0.225 \\
         Baseline$\mathit{+RPE(TD)}$ & 0.207 & 0.231 & 0.143 & 0.188 & Baseline$\mathit{+RPE(TD)+Drop}$ & 0.203 & 0.227 & 0.146 & 0.192 \\
         Baseline$\mathit{+RPE(T1)}$ & 0.187 & 0.209 & 0.164 & 0.216 & Baseline$\mathit{+RPE(T1)+Drop}$ & 0.176 & 0.196 & 0.166 & 0.218 \\
         Baseline$\mathit{+LPE(TD)}$ & 0.190 & 0.212 & 0.136 & 0.179 & Baseline$\mathit{+LPE(TD)+Drop}$ & 0.202 & 0.226 & 0.151 & 0.199 \\
         Baseline$\mathit{+LPE(T1)}$ & 0.184 & 0.205 & 0.173 & 0.227 & Baseline$\mathit{+LPE(T1)+Drop}$ & 0.178 & 0.199 & 0.177 & 0.233 \\
         Baseline$\mathit{+CPE(TD)}$ & 0.141 & 0.157 & 0.157 & 0.206 & Baseline$\mathit{+CPE(TD)+Drop}$ & 0.123 & 0.136 & 0.145 & 0.191 \\
         Baseline$\mathit{+CPE(T1)}$ & 0.111 & 0.123 & 0.069 & 0.090 & Baseline$\mathit{+CPE(T1)+Drop}$ & 0.129 & 0.144 & 0.090 & 0.118 \\
      \end{tabular}
    }
    \caption{The results of the model without softmax as the baseline. $\mathit{Att}$ applies key and value embedding with scaling factor for the size of frames without softmax (\Cref{tab:tab8-a}).}
    \label{tab:tab10-a}
  \end{subtable}
  \begin{subtable}[c]{\textwidth}
    \centering
    \resizebox{0.99\textwidth}{!}{
      \begin{tabular}{l|cc|cc|l|cc|cc}
        \noalign{\smallskip}\noalign{\smallskip}
        \hline
        \multirow{2}{*}{Setting} & \multicolumn{2}{c|}{\datasetfirst} & \multicolumn{2}{c|}{\datasetsecond} & \multirow{2}{*}{Setting} & \multicolumn{2}{c|}{\datasetfirst} & \multicolumn{2}{c}{\datasetsecond} \\
          & $\mathit{\tau}$ & $\mathit{\rho}$ & $\mathit{\tau}$ & $\mathit{\rho}$ & & $\mathit{\tau}$ & $\mathit{\rho}$ & $\mathit{\tau}$ & $\mathit{\rho}$ \\
        \hline
         Baseline($\mathit{Att_{T\&D}}$) & 0.207 & 0.231 & 0.193 & 0.253 & Baseline$\mathit{+Drop}$ & 0.204 & 0.228 & 0.193 & 0.253 \\
         Baseline$\mathit{+FPE(TD)}$ & 0.222 & 0.248 & 0.188 & 0.247 & \textbf{Baseline}$\mathit{\textbf{+FPE(TD)+Drop}}$ & \textbf{0.225} & \textbf{0.251} & \textbf{0.191} & \textbf{0.251} \\
         Baseline$\mathit{+FPE(T1)}$ & 0.207 & 0.231 & 0.183 & 0.241 & Baseline$\mathit{+FPE(T1)+Drop}$ & 0.199 & 0.222 & 0.184 & 0.243 \\
         Baseline$\mathit{+RPE(TD)}$ & 0.200 & 0.223 & 0.189 & 0.248 & Baseline$\mathit{+RPE(TD)+Drop}$ & 0.205 & 0.229 & 0.189 & 0.248 \\
         Baseline$\mathit{+RPE(T1)}$ & 0.214 & 0.239 & 0.184 & 0.242 & Baseline$\mathit{+RPE(T1)+Drop}$ & 0.209 & 0.233 & 0.184 & 0.243 \\
         Baseline$\mathit{+LPE(TD)}$ & 0.216 & 0.241 & 0.185 & 0.243 & Baseline$\mathit{+LPE(TD)+Drop}$ & 0.217 & 0.242 & 0.186 & 0.245 \\
         Baseline$\mathit{+LPE(T1)}$ & 0.197 & 0.220 & 0.181 & 0.239 & Baseline$\mathit{+LPE(T1)+Drop}$ & 0.207 & 0.231 & 0.181 & 0.238 \\
         Baseline$\mathit{+CPE(TD)}$ & 0.204 & 0.228 & 0.187 & 0.246 & Baseline$\mathit{+CPE(TD)+Drop}$ & 0.202 & 0.226 & 0.190 & 0.249 \\
         Baseline$\mathit{+CPE(T1)}$ & 0.209 & 0.233 & 0.192 & 0.253 & Baseline$\mathit{+CPE(T1)+Drop}$ & 0.205 & 0.228 & 0.190 & 0.250 \\
      \end{tabular}
    }
    \caption{The results of the model with softmax as the baseline. $\mathit{Att_{T\&D}}$ applies key and value embedding with softmax (\Cref{tab:tab8-b}).}
    \label{tab:tab10-b}
  \end{subtable}
  \caption{The ablation study for methods in PGL-SUM. $\mathit{FPE}$ is fixed positional encoding, $\mathit{RPE}$ is relative positional encoding, $\mathit{LPE}$ is learnable positional encoding, and $\mathit{CPE}$ is conditional positional encoding. $\mathit{T}$ is the frame axis, and $\mathit{TD}$ is the frame and dimension axes for positional encoding. $\mathit{Drop}$ exploits dropout to features after positional encoding.}
  \label{tab:tab10}
\end{table}

Unlike existing transformers, PGL-SUM \cite{pglsum} proves effects when applying positional encodings and dropouts to the multiplication outputs between key and query vectors. We adopted the same methods to further improve the model’s positional recognition effectiveness by adding the positional encodings and applying dropouts to CNN's outputs. We use 0.5 for the dropout ratio, the same as \cite{pglsum}.\par

The best performance for the model without and with softmax is achieved by Baseline$\mathit{+Drop}$ and Baseline$\mathit{+FPE(TD)+Drop}$, respectively, as shown in \Cref{tab:tab10}. In \Cref{tab:tab10-b}, some models perform slightly better than Baseline$\mathit{+FPE(TD)+Drop}$ on \datasetsecond, but their performance is considerably worse than the selected one on \datasetfirst. Comparing the best performance in both tables, we observe that the performance of the model with softmax (Baseline$\mathit{+FPE(TD)+Drop}$) is better than that without softmax (Baseline$\mathit{+Drop}$) for all datasets. Although their performance is similar on \datasetsecond, the baseline model without softmax shows 0.219 and 0.244 for Kendall's and Spearman's coefficients, respectively, on \datasetfirst. The baseline model with softmax shows 0.225 and 0.251 for Kendall's and Spearman's coefficients, respectively, on \datasetfirst. Based on the performance, we chose the model with softmax using $\mathit{FPE(TD)}$ and $\mathit{Drop}$ as the final model.

\subsection{CLS token}
\label{sec:sup_cls_ext}

\begin{table}[ht]
  \centering
  \resizebox{0.5\textwidth}{!}{
    \begin{tabular}{l|cc|cc}
      \noalign{\smallskip}\noalign{\smallskip}
      \hline
      \multirow{2}{*}{Setting} & \multicolumn{2}{c|}{\datasetfirst} & \multicolumn{2}{c}{\datasetsecond} \\
        & $\mathit{\tau}$ & $\mathit{\rho}$ & $\mathit{\tau}$ & $\mathit{\rho}$ \\
      \hline
       Baseline($\mathit{Att_{T\&D}}$) & 0.225 & 0.251 & 0.191 & 0.251 \\
       Baseline$\mathit{+CLS_{CNN}}$ & 0.232 & 0.259 & 0.194 & 0.254 \\
       Baseline$\mathit{+CLS_{SM}}$ & 0.233 & 0.260 & 0.193 & 0.254 \\
       \textbf{Baseline}$\mathit{\textbf{+CLS}_{\textbf{Final}}}$ & \textbf{0.236} & \textbf{0.263} & \textbf{0.194} & \textbf{0.254} \\
    \end{tabular}
  }
  \caption{The ablation study for the CLS token. $\mathit{Att_{T\&D}}$ is the baseline model applying FPE and dropout (\Cref{tab:tab10}). $\mathit{CLS}$ is the model that uses the CLS token and combines it with frame features after CNN ($\mathit{CNN}$) or softmax ($\mathit{SM}$) or creating final features (Final).}
  \label{tab:tab11}
\end{table} 

We further test the CLS token \cite{ViT} at different combining places. \modelname\, fuses the CLS token with input frame features right after employing CNN or softmax or creating final features. The final features are created by applying attention maps to input features.\par

Combining the CLS token after creating the final features yields the best performance, as shown in \Cref{tab:tab11}. We hypothesize that the reason is that the classifier is generalized. The CLS token is trained jointly with the model to reflect the overall information of the dataset. The global cues of the dataset generalize the classifier because fully connected layers are found in the classifier. Thus, all results using the CLS token improved the baseline. Moreover, adding the CLS token after creating the final features means incorporating the CLS token just before the classifier. For this reason, the best performance is achieved by delivering the CLS token without changes and generalizing the classifier the most.

\subsection{Skip connection}
\label{sec:sup_sc_ext}

\begin{table}[ht]
  \centering
  \resizebox{\textwidth}{!}{
    \begin{tabular}{l|cc|cc|l|cc|cc}
      \noalign{\smallskip}\noalign{\smallskip}
      \hline
      \multirow{2}{*}{Setting} & \multicolumn{2}{c|}{\datasetfirst} & \multicolumn{2}{c|}{\datasetsecond} & \multirow{2}{*}{Setting} & \multicolumn{2}{c|}{\datasetfirst} & \multicolumn{2}{c}{\datasetsecond} \\
      & $\mathit{\tau}$ & $\mathit{\rho}$ & $\mathit{\tau}$ & $\mathit{\rho}$ & & $\mathit{\tau}$ & $\mathit{\rho}$ & $\mathit{\tau}$ & $\mathit{\rho}$ \\
      \hline
       Baseline($\mathit{Att_{T\&D}}$) & 0.236 & 0.263 & 0.194 & 0.254 & & & & & \\
       Baseline$\mathit{+SC_{KC}}$ & 0.233 & 0.261 & 0.193 & 0.253 & \textbf{Baseline}$\mathit{\textbf{+SC}_{\textbf{KC}}\textbf{+LN}}$ & \textbf{0.243} & \textbf{0.271} & \textbf{0.192} & \textbf{0.252} \\
       Baseline$\mathit{+SC_{CF}}$ & 0.115 & 0.128 & 0.043 & 0.056 & Baseline$\mathit{+SC_{CF}+LN}$ & 0.162 & 0.181 & 0.052 & 0.068 \\
       Baseline$\mathit{+SC_{IF}}$ & 0.126 & 0.141 & -0.018 & -0.024 & Baseline$\mathit{+SC_{IF}+LN}$ & 0.163 & 0.181 & 0.186 & 0.244 \\
    \end{tabular}
  }
  \caption{The ablation study for the skip connection. $\mathit{Att_{T\&D}}$ is the baseline model incorporating the CLS token after creating the final features (\Cref{tab:tab11}). $\mathit{SC}$ means skip connection. $\mathit{KC}$ is the key embedding output fused with CNN output. $\mathit{CF}$ is CNN output combined with final features. $\mathit{IF}$ is the combined input and final features. $\mathit{LN}$ is layer normalization exploited immediately after the skip connection.}
  \label{tab:tab12}
\end{table}

We finally verify the skip connection \cite{resnet} for stable optimization of \modelname, as shown in \Cref{tab:tab12}. Without layer normalization, adopting the skip connection by adding outputs from the key embedding and CNN ($\mathit{SC_{KC}}$) yields the best performance among all settings. However, it yields slightly worse performance than the baseline model for all datasets. Using layer normalization with $\mathit{SC_{KC}}$ ($\mathit{SC_{KC}+LN}$) shows slightly less performance than the baseline model on \datasetsecond, whereas it yields much better performance than the baseline on \datasetfirst. For better overall performance, we selected the combination of skip connections, which is the sum of both key embedding and CNN outputs, and layer normalization as the final model.

%%%%%%%%%%%%%%%%%%%%%%%%%%%%%%%%%%%%%%%%%%%%%%%%%%%%%%%%%%%%%%%%%%%%%%%%%%%%%%%%%%%%%%%%%%%%%%%%%%
%%%%%%%%%%%%%%%%%%%%%%%%%%%%%%%%%%%%%%%%%%%%%%%%%%%%%%%%%%%%%%%%%%%%%%%%%%%%%%%%%%%%%%%%%%%%%%%%%%
%%%%%%%%%%%%%%%%%%%%%%%%%%%%%%%%%%%%%%%%%%%%%%%%%%%%%%%%%%%%%%%%%%%%%%%%%%%%%%%%%%%%%%%%%%%%%%%%%%

\section{Hyperparameter setting}
\label{sec:sup_hyperparameter_setting}

\begin{table}[ht]
  \centering
  \begin{subtable}[c]{\textwidth}
    \centering
    \resizebox{0.9\textwidth}{!}{
      \begin{tabular}{l|l|ccccc}
        \hline
        Dataset & Correlation & \textbf{Batch size=1} & Batch size=25\% & Batch size=50\% & Batch size=75\% & Batch size=100\% \\
        \hline
         \datasetfirst & $\mathit{\tau}$ & \textbf{0.243} & 0.214 & 0.210 & 0.211 & 0.202 \\
          & $\mathit{\rho}$ & \textbf{0.271} & 0.239 & 0.235 & 0.235 & 0.225 \\
        \hline
         \datasetsecond & $\mathit{\tau}$ & \textbf{0.192} & 0.173 & 0.161 & 0.163 & 0.152 \\
          & $\mathit{\rho}$ & \textbf{0.252} & 0.228 & 0.212 & 0.214 & 0.200 \\
      \end{tabular}
    }
    \caption{The ablation study for different batch sizes. Each percentage is the batch size ratio of the entire training dataset.}
    \label{tab:tab13-a}
  \end{subtable}
  \begin{subtable}[c]{\textwidth}
    \centering
    \resizebox{0.8\textwidth}{!}{
      \begin{tabular}{l|l|ccccc}
        \hline
        Dataset & Correlation & Dropout=0.9 & Dropout=0.8 & Dropout=0.7 & \textbf{Dropout=0.6} & Dropout=0.5 \\
        \hline
        \datasetfirst & $\mathit{\tau}$ & 0.227 & 0.237 & 0.240 & \textbf{0.246} & 0.243 \\
         & $\mathit{\rho}$ & 0.253 & 0.264 & 0.268 & \textbf{0.275} & 0.271 \\
        \datasetsecond & $\mathit{\tau}$ & 0.198 & 0.196 & 0.194 & \textbf{0.192} & 0.192 \\
         & $\mathit{\rho}$ & 0.260 & 0.258 & 0.255 & \textbf{0.252} & 0.252 \\
        \cline{1-6}
        Dataset & Correlation & Dropout=0.4 & Dropout=0.3 & Dropout=0.2 & Dropout=0.1 & \\
        \cline{1-6}
         \datasetfirst & $\mathit{\tau}$ & 0.243 & 0.239 & 0.241 & 0.237 & \\
         & $\mathit{\rho}$ & 0.271 & 0.267 & 0.269 & 0.264 & \\
         \datasetsecond & $\mathit{\tau}$ & 0.192 & 0.192 & 0.190 & 0.192 & \\
         & $\mathit{\rho}$ & 0.252 & 0.253 & 0.250 & 0.252 & \\
      \end{tabular}
    }
    \caption{The ablation study for different dropout ratios used for CNN outputs with a single batch size (\Cref{tab:tab13-a}).}
    \label{tab:tab13-b}
  \end{subtable}
  \begin{subtable}[c]{\textwidth}
    \centering
    \resizebox{0.65\textwidth}{!}{
      \begin{tabular}{l|l|ccccc}
        \hline
        Dataset & Correlation & WD=1e-1 & WD=1e-2 & WD=1e-3 & WD=1e-4 & WD=1e-5 \\
        \hline
        \datasetfirst & $\mathit{\tau}$ & 0.174 & 0.178 & 0.176 & 0.203 & 0.227 \\
         & $\mathit{\rho}$ & 0.194 & 0.199 & 0.197 & 0.226 & 0.253 \\
        \datasetsecond & $\mathit{\tau}$ & 0.055 & 0.082 & 0.195 & 0.195 & 0.199 \\
         & $\mathit{\rho}$ & 0.070 & 0.107 & 0.256 & 0.255 & 0.261 \\
        \cline{1-6}
        Dataset & Correlation & WD=1e-6 & \textbf{WD=1e-7} & WD=1e-8 & WD=0 & \\
        \cline{1-6}
         \datasetfirst & $\mathit{\tau}$ & 0.237 & \textbf{0.246} & 0.242 & 0.246 & \\
         & $\mathit{\rho}$ & 0.264 & \textbf{0.274} & 0.270 & 0.275 & \\
         \datasetsecond & $\mathit{\tau}$ & 0.194 & \textbf{0.194} & 0.193 & 0.192 & \\
         & $\mathit{\rho}$ & 0.255 & \textbf{0.255} & 0.253 & 0.252 & \\
      \end{tabular}
    }
    \caption{The ablation study for different weight decays with a single batch size and dropout ratio of 0.6 (\Cref{tab:tab13-b}). WD means weight decay.}
    \label{tab:tab13-c}
  \end{subtable}
  \begin{subtable}[c]{\textwidth}
    \centering
    \resizebox{0.55\textwidth}{!}{
      \begin{tabular}{l|l|cccc}
        \hline
        Dataset & Correlation & LR=1e-1 & LR=1e-2 & \textbf{LR=1e-3} & LR=1e-4 \\
        \hline
        \datasetfirst & $\mathit{\tau}$ & -0.292 & 0.164 & \textbf{0.246} & 0.211 \\
         & $\mathit{\rho}$ & -0.284 & 0.182 & \textbf{0.274} & 0.236 \\
        \datasetsecond & $\mathit{\tau}$ & -0.089 & 0.060 & \textbf{0.194} & 0.162 \\
         & $\mathit{\rho}$ & -0.080 & 0.080 & \textbf{0.255} & 0.213 \\
        \cline{1-6}
        Dataset & Correlation & LR=1e-5 & LR=1e-6 & LR=1e-7 & LR=1e-8 \\
        \cline{1-6}
         \datasetfirst & $\mathit{\tau}$ & 0.162 & 0.163 & 0.168 & 0.153 \\
         & $\mathit{\rho}$ & 0.181 & 0.182 & 0.187 & 0.171 \\
         \datasetsecond & $\mathit{\tau}$ & 0.101 & 0.030 & 0.035 & 0.037 \\
         & $\mathit{\rho}$ & 0.133 & 0.039 & 0.046 & 0.049 \\
      \end{tabular}
    }
    \caption{The ablation study for different learning rates with a single batch size, a dropout ratio of 0.6, and weight decay of 1e-7 (\Cref{tab:tab13-c}). LR means learning rate.}
    \label{tab:tab13-d}
  \end{subtable}
  \caption{The ablation study for different hyperparameter settings.}
  \label{tab:tab13}
\end{table}

Here, we test the model with different hyperparameter values. The best performance of the model is achieved with a single batch size, and it keeps decreasing with larger batch sizes, as shown in \Cref{tab:tab13-a}. Thus, we use the single batch size.\par

The bigger the dropout ratio, the better the model's performance on \datasetsecond, as shown in \Cref{tab:tab13-b}. However, the performance of the model is bad if the dropout ratio is too large, so we chose 0.6 as the dropout ratio, considering both performance on \datasetfirst\, and \datasetsecond.\par

We fixed the dropout ratio at 0.6 and tested with various values of weight decay, as shown in \Cref{tab:tab13-c}. The performance increased as the value of weight decay decreased. When weight decay is 1e-7, the performance on \datasetfirst\, is similar to that without weight decay. However, it shows slightly better performance on \datasetsecond\, than the model with 0 weight decay. Thus, considering the overall performance on \datasetfirst\, and \datasetsecond, we select 1e-7 as the final value for weight decay.\par

In \Cref{tab:tab13-d}, we finally test different learning rates with 1e-7 as weight decay. When the learning rate is too large, the performance is terrible for all datasets. When the learning rate is too small, the performance is also poor. When the learning rate is 1e-3, it shows the best performance, so we decided on this value as the final learning rate.

\end{document}